\RequirePackage{letltxmacro}
\LetLtxMacro{\LaTeXtextbf}{\textbf}
\documentclass{ieeeaccess}
\LetLtxMacro{\textbf}{\LaTeXtextbf}

\IEEEoverridecommandlockouts

\usepackage{cite}
\usepackage{amsmath,amssymb,amsfonts}
\usepackage[ruled]{algorithm2e}
\usepackage{graphicx}
\usepackage{textcomp}

\usepackage{booktabs}
\usepackage{multirow}
\usepackage{hyperref}
\usepackage{cleveref}
\definecolor{cvprblue}{rgb}{0.21,0.49,0.74}
\makeatletter

\DeclareRobustCommand\onedot{\futurelet\@let@token\@onedot}
\def\@onedot{\ifx\@let@token.\else.\null\fi\xspace}
\def\eg{\emph{e.g}\onedot} 

\def\ie{\emph{i.e}\onedot}

\def\etal{\emph{et al}\onedot}

\crefname{figure}{fig.}{figs.}
\Crefname{Figure}{Fig.}{Figs.}

\makeatother

\usepackage{bm}
\makeatletter
\AtBeginDocument{\DeclareMathVersion{bold}
\SetSymbolFont{operators}{bold}{T1}{times}{b}{n}
\SetSymbolFont{NewLetters}{bold}{T1}{times}{b}{it}
\SetMathAlphabet{\mathrm}{bold}{T1}{times}{b}{n}
\SetMathAlphabet{\mathit}{bold}{T1}{times}{b}{it}
\SetMathAlphabet{\mathbf}{bold}{T1}{times}{b}{n}
\SetMathAlphabet{\mathtt}{bold}{OT1}{pcr}{b}{n}
\SetSymbolFont{symbols}{bold}{OMS}{cmsy}{b}{n}
\renewcommand\boldmath{\@nomath\boldmath\mathversion{bold}}}
\makeatother

\def\BibTeX{{\rm B\kern-.05em{\sc i\kern-.025em b}\kern-.08em
    T\kern-.1667em\lower.7ex\hbox{E}\kern-.125emX}}

\begin{document}
\history{Date of publication xxxx 00, 0000, date of current version xxxx 00, 0000.}
\doi{10.1109/ACCESS.2024.0429000}

\title{Dynamic Test-Time Augmentation via Differentiable Functions}
\author{\uppercase{Shohei Enomoto}\authorrefmark{1}, 
\uppercase{Monikka Roslianna Busto}\authorrefmark{1}, and \uppercase{Takeharu Eda}\authorrefmark{1},
\IEEEmembership{Member, IEEE}}

\address[1]{NTT, Musashino, Tokyo 180-8585, Japan}

\tfootnote{}

\markboth
{S. Enomoto \headeretal: Dynamic Test-Time Augmentation via Differentiable Functions}
{S. Enomoto \headeretal: Dynamic Test-Time Augmentation via Differentiable Functions}

\corresp{Corresponding author: Shohei Enomoto (e-mail: alesana882@gmail.com).}

\begin{abstract}
Distribution shifts, which often occur in the real world, degrade the accuracy of deep learning systems, and thus improving robustness to distribution shifts is essential for practical applications.
To improve robustness, we study an image enhancement method that generates recognition-friendly images without retraining the recognition model.
We propose a novel image enhancement method, DynTTA, which is based on differentiable data augmentation techniques and generates a blended image from many augmented images to improve the recognition accuracy under distribution shifts.
In addition to standard data augmentations, DynTTA also incorporates deep neural network-based image transformation, further improving the robustness.
Because DynTTA is composed of differentiable functions, it can be directly trained with the classification loss of the recognition model.
In experiments with widely used image recognition datasets using various classification models, DynTTA improves the robustness with almost no reduction in classification accuracy for clean images, thus outperforming the existing methods.
Furthermore, the results show that robustness is significantly improved by estimating the training-time augmentations for distribution-shifted datasets using DynTTA and retraining the recognition model with the estimated augmentations.
DynTTA is a promising approach for applications that require both clean accuracy and robustness.
Our code is available at \url{https://github.com/s-enmt/DynTTA}.
\end{abstract}


\titlepgskip=-21pt

\maketitle
\section{Introduction}

With the development of deep learning, the field of visual recognition has made great progress, and services using deep learning are becoming more practical.
Many deep learning models are trained and validated on images from the same distribution.
However, images in the real world are subject to distribution shifts due to various factors such as weather conditions, sensor noise, blurring, and compression artifacts. 
Deep learning models do not usually take these distribution shifts into account, causing the accuracy to reduce in the presence of such artifacts.
This is fatal for safety-critical applications, such as automated driving, where the environment changes frequently.

There are two main approaches to solving this problem: training robust recognition models and image enhancement.
The former uses data augmentation~\cite{augmix,pixmix,ipmix}, or training algorithms such that the recognition model has high robustness.
The latter uses test-time augmentation~\cite{lp} or deep neural networks~(DNNs)~\cite{urie} to transform the distorted images into recognition-friendly images that are easily recognized by the recognition model.
These approaches can be used in combination to achieve higher robustness.
This paper focuses on the image enhancement approach, which is highly practical because it is used before inference by the pretrained recognition model without retraining the recognition model.
We found that DNN-based image enhancement~\cite{urie} overfits to particular transformation patterns, resulting in excessive image transformation even for clean images and reduced clean accuracy.
Test-time augmentation-based image enhancement~\cite{lp}, which dynamically selects the best of several data augmentations for an input image, improves robustness while maintaining clean accuracy.
However, its augmentation search space is limited, so the improvement in robustness is small.

\begin{figure*}[tb!]
\centering
\includegraphics[width=0.7\linewidth]{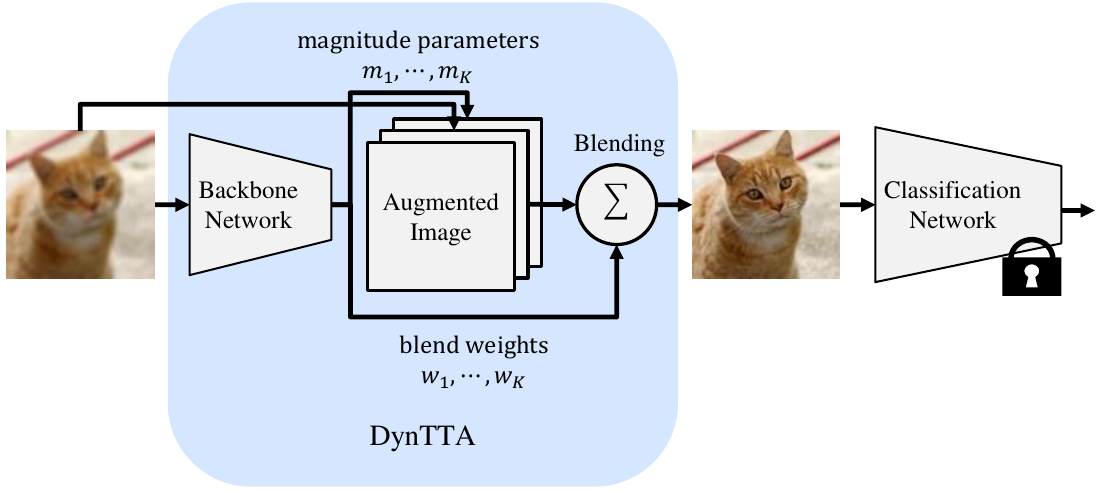}
\caption{
DynTTA is used before inference by the classification network to generate recognition-friendly images. 
First, DynTTA outputs the magnitude parameters and blend weights. 
Next, predefined data augmentations are performed using the magnitude parameters. 
Finally, the output image is generated by linearly combining the augmented images with the blend weights. The generated images are input to the classification network. 
In this paper, these image transformation models that are used before the classification network are referred to as enhancement models.
}
\label{fig:DynTTA_overview}
\end{figure*}

To remove these limitations, we propose a novel image enhancement method, DynTTA.
DynTTA uses differentiable data augmentation techniques~\cite{kornia,dda,dada} and image blending~\cite{augmix,augmax,pixmix,ipmix} to dynamically generate the recognition-friendly image for an input image from a huge augmentation space.
DynTTA incorporates a DNN-based image transformation by considering it as a data augmentation.
Data augmentations work as a hint for learning diverse image transformations and avoid overfitting, resulting in significantly improving robustness without losing clean accuracy.
An overview diagram of DynTTA is shown in \Cref{fig:DynTTA_overview}.
DynTTA takes an image as input and outputs the magnitude parameters and blend weights for predefined data augmentation.
Each augmentation is performed on the basis of the magnitude parameters.
These augmented images are linearly combined with blend weights to generate the output image.
The recognition model takes DynTTA's output images as input and performs inference.
Because DynTTA is composed of differentiable functions, it can be directly trained with the loss of the recognition task.
As a result, DynTTA transforms distorted images into recognizable ones to improve accuracy without needing model retraining.
In addition, we hypothesize that training with the inverse operations of highly weighted augmentations by DynTTA would make the classification model robust against the given distribution. 
We thus propose a novel method using DynTTA: estimating effective training-time augmentations and retraining classification models with those augmentations.

DynTTA was evaluated on widely used image classification datasets and various classification models.
We introduce a practical training and evaluation scenario, the blind setting, which does not assume the type of test-time distribution shifts and has not been used in existing literature.
DynTTA improves the accuracy for distorted images while maintaining the accuracy for clean images, which is a better result than those of existing methods.
Specifically, experimental results on the ImageNet dataset~\cite{imagenet} show that DynTTA improves the robustness of pre-trained ResNet50~\cite{resnet} by 5.30 percentage points while maintaining clean accuracy.
Moreover, retraining with the estimated training-time augmentation on the PACS dataset~\cite{pacs} improves the accuracy of ResNet50 by 6.43 percentage points.
DynTTA is a promising approach for applications that require both clean accuracy and robustness, such as automated driving.

The main contributions of this paper are as follows.
\begin{itemize}
\item We propose DynTTA, a novel image enhancement method based on differentiable data augmentation techniques and image blending. 
DynTTA generates diverse images by transformations with various data augmentations, including a DNN, which significantly improves robustness while maintaining clean accuracy.
\item We propose a novel method to estimate effective training-time augmentations and show that retraining the classification model with the estimated augmentations significantly improves accuracy.
\item Our extensive experiments show that DynTTA improves robustness compared to existing methods while maintaining clean accuracy.
In addition, experimental results in the blind setting show that DynTTA is practical under various distribution shifts.
\end{itemize}

\section{Related Work}

\subsection{Robustness of DNNs}
DNNs are known to be vulnerable to image distortions, which often occur in the real world.
These image distortions have been studied in several works.
Diamond~\etal~\cite{diamond2021dirty} showed that defects such as noise and blur in real-world sensors degrade performance in image recognition networks.
Pei~\etal~\cite{pei2018does} found that haze reduces the accuracy of image classification and experimented that the dehaze method for human visibility does not effectively improve image classification performance.
Pei~\etal~\cite{8889765} empirically studied real-world image degradation problems for nine kinds of degraded images:-— hazy, motion-blurred, fish-eye, underwater, low resolution, salt-and-peppered, white Gaussian noise, Gaussian-blurred, and out-of-focus.
Various datasets have been proposed to evaluate the robustness of DNNs.
Geirhos~\etal~\cite{stylized-IN} made a style-transformed dataset to demonstrate that DNNs recognize objects by their texture rather than their shape.
Hendrycks and Dietterich~\cite{IN-C} proposed 19 algorithmically generated corruption datasets in five levels, belonging to four categories: noise, blur, weather, and digital.
They showed that DNNs trained on a clean dataset have lower accuracy on these corruption datasets.
In addition, they introduced a dataset of naturally occurring adversarial examples in the real world~\cite{IN-A} and a dataset containing distribution shifts that occur in the real world, such as image style, image blurriness, geographic location, camera operation, and more~\cite{IN-R}.
In this study, we evaluate our method on these datasets.

\subsection{Training Robust Classification Models}
Many works have been studied to improve the robustness of DNNs to naturally occurring image distortions.
One way to improve the robustness is the data augmentation.
Mixing multiple data augmentations~\cite{augmix,augmax,noisymix,pixmix,ipmix,vaish2024fourier} and adding noise during training~\cite{rusak2020simple,tsiligkaridis2021misclassification} have been proposed as effective data augmentations to improve robustness.
Mintun~\etal~\cite{mintun2021interaction} found that the similarity between data augmentation and test-time image corruption is strongly correlated with performance.
Training algorithms to improve robustness have also been studied.
Disentangled learning via auxiliary batch normalization~\cite{xie2020adversarial,mixprop,fast_advprop,entprop}, a method to improve adversarial training, improves not only robustness but also clean accuracy.
Li~\etal~\cite{li2020shape} proposed an algorithm to train the model paying complementary attention to the shape and texture of the objects.
Since these methods are used when training recognition models, they can be used in combination with image enhancement methods, which assume that the recognition models are pretrained.
Our experimental results show that combining the recognition models pretrained by these methods with image enhancement further improves the robustness.

\subsection{Image Enhancement}
This section presents two image enhancement approaches, test-time augmentation and image transformation.
\Cref{tab:compare_approaches} shows a summary of the pros and cons of these approaches and DynTTA.

\begin{table}[tb!]
\centering
\caption{
Pros and cons of each approach.
DynTTA achieves both maintaining clean accuracy and improving robustness.
}
\label{tab:compare_approaches}
\begin{tabular}{l|cc}
\toprule
                       & Clean Accuracy & Robustness \\ \midrule
Test-time Augmentation & $\checkmark$       & $\times$         \\
Image Transformation   & $\times$        & $\checkmark$      \\
DynTTA                 & $\checkmark$       & $\checkmark$      \\ \bottomrule
\end{tabular}
\end{table}

\subsubsection{Test-time Augmentation}
Recognition accuracy can be improved by using data augmentation during not only the training time but also the test time.
Various test-time augmentation methods have been proposed, using simple geometric transformations such as flip and crop~\cite{perez2021enhancing,yang2018deep,8978185,alexnet,googlenet,resnet}, mixup~\cite{lee2021test,mixup_inf}, and augmentation in the embedding space~\cite{ashukha2021mean}.
Shanmugam~\etal~\cite{shanmugam2021better} experimentally analyzed why and when test-time augmentation works, and Kimura~\etal~\cite{understanding_TTA} gave theoretical guarantees.
These studies use only specific data augmentations for all test set images.
Dynamically selecting the best data augmentation for each input image is expected to improve the accuracy. 
Lyzhov~\etal~\cite{gps} proposed a greedy search-based test-time augmentation method to find the augmentation policy on the test set, but it is not optimal for each input image.
Kim~\etal~\cite{lp} proposed a module called Loss Predictor to predict the classification loss of augmented images, allowing dynamic selection of the best one for the input image from 12 data augmentations.
However, despite the augmentation space actually being infinite, these methods have significant limitations on the augmentation space due to computational cost and the improvement in robustness is small.
Our method eliminates this limitation and achieves better robustness than existing methods by generating the best augmented image from a large augmentation space.
In addition, we propose the training-time augmentation estimation method for distribution-shifted datasets using test-time augmentation and show that retraining the classification model by estimated augmentations significantly improves accuracy.

\subsubsection{Image Transformation}
Several methods have been proposed to improve the recognition accuracy by image transformation using DNNs.
Sharma~\etal~\cite{8578522} proposed convolutional neural network-based enhancement filters that enhance image-specific details to improve recognition accuracy.
Subsequent work~\cite{liu2018when,Okatani2019ImprovingGA,urie,lee2021task,bts} used DNN to transform corrupted images into recognition-friendly ones.
Milanfar~\etal~\cite{talebi2021learning} showed that not only image enhancement but also resizing at the same time improves the recognition accuracy.
These methods use DNNs to transform images, but tend to perform excessive transformations to remove distortions, which reduces the accuracy of clean images.
By considering these image enhancement methods as a kind of data augmentation, we treat them as part of DynTTA.
DNN-based transformations reduce clean accuracy because they overfit particular transformation patterns that removes distortions.
We integrate DNN transformation and data augmentations to enable learning of diverse image transformations, which avoids overfitting and improves robustness while maintaining clean accuracy.

\section{DynTTA}
\label{sec:dyntta}
\subsection{Key Ideas}
Many data augmentations have continuous magnitude parameters (\eg, rotation has a continuous magnitude parameter from 0 to 360 degrees).
In addition, there are many combinations of two or more data augmentations (\eg, augmentation is performed in order of rotation, contrast, and sharpness).
These facts make the augmentation space infinite.
To solve the test-time augmentation challenge of selecting the best combination of data augmentations and their optimal magnitude parameters for the test image from an infinite space, we introduce two key ideas.
The first is to find the local optimal magnitude parameter by optimization.
Since most data augmentations are differentiable with respect to the magnitude parameter~\cite{kornia,dda,dada,augnet}, the local optimal magnitude parameter is obtained by optimization algorithms, such as gradient descent.
The second is to represent the combination of augmentation with image blending, which is the process of affine combination of multiple images.
For example, if the weights are $[1,0,0]$, only the 1st augmented image is selected, and if the weights are $[0.5,0,0.5]$, the 1st and 3rd augmented images are combined.
Note that image blending does not provide the performing order of the augmentation.
We call these ideas ``Optimization’’ and ``Blending’’, respectively.
On the basis of these key ideas, we propose a novel image enhancement method, DynTTA.

\subsection{Overview of DynTTA}
The overview of DynTTA is shown in \Cref{fig:DynTTA_overview}.
The differentiable data augmentations used in DynTTA and their range of magnitude parameters $M$ are defined.
Most data augmentations, such as rotation, have a magnitude parameter.
For data augmentations that do not have a magnitude parameter, such as auto-contrast, $M$ is not defined.
\Cref{sec:aug_space} details the data augmentations of DynTTA.

Any DNN can be used as the backbone of DynTTA by modifying its output layer.
DynTTA takes an image as input and outputs the magnitude parameters $m_1,\cdots,m_{K}$ and blend weights $w_1,\cdots,w_{K}$.
$K$ is the number of data augmentations.
The magnitude parameter is mapped to a value in the range $(0,1)$ or $(-1,1)$ using an activation function $f_k^{\mathrm{act}}$ such as sigmoid function or hyperbolic tangent function.
It is then multiplied by the predefined magnitude range $M$.
The blend weights are converted to weights that sum to 1 by the softmax function.
$m$ and $w$ are shown in the following equation.
\begin{align}
\hat{m_k} &= \left\{ \begin{array}{ll}
M_k f_k^{\mathrm{act}}~(m_k) \; & (\text{if $M_k$ and $f_k^{\mathrm{act}}$ are defined}) \\
\emptyset \; & (\text{else}), \\
\end{array} \right. \\
\hat{w_k} &= \frac{\exp(w_k)}{\sum_{i=1}^{K} \exp(w_i)}.
\end{align}

By using the magnitude parameters and blend weights, DynTTA generates a recognition-friendly image.
$K$ data augmentations are performed with magnitude parameters $\hat{m}$, then DynTTA generates a blended image by linearly combining augmented images with $\hat{w}$.
DynTTA outputs the image using the following equation:
\begin{align}
\hat{x} &= \sum^{K}_{k=1} \hat{w_k} O_{k}~(x, \hat{m_k}),
\end{align}
where $x$ is the input image, $O_k$ is the $k$-th augmentation operation, and $\hat{x}$ is the output image.

\subsection{Augmentation Space}
\label{sec:aug_space}
The data augmentations and their $f_k^{\mathrm{act}}$ and $M$ used by DynTTA are shown in \Cref{tab:augmentation_detail}.
Because DynTTA obtains magnitude parameters through optimization, only differentiable data augmentations are used.
This study uses the kornia library~\cite{kornia} and deals only with general data augmentation.
We do not employ data augmentation such as shear because we believe it is ineffective when used during testing.
Flip is a typical test-time augmentation, but it is not used because it is non-differentiable.
Placing low-pass and high-pass filters within the DNN architecture~\cite{zou2020delving,rao2021global,vasconcelos2021impact,9455411} or using them as training-time augmentation~\cite{hossain2019distortion,yin2019fourier,soklaski2021fourier} is known to improve robustness.
Inspired by these studies, we try to improve the robustness by using low-pass and high-pass filters as test-time augmentation.
These filters have a filter size parameter that is not obtained via gradient descent.
Therefore, this parameter is discretized, and multiple filters are prepared.
19 low-pass and 19 high-pass filters are made, each with a filter size from $0.05$ to $0.95$ in increments of $0.05$.
URIE uses a DNN to generate recognition-friendly images.
Such an image transformation model is applied to DynTTA by considering it as a kind of augmentation.
In this case, the image transformation model is trained simultaneously with DynTTA.

DynTTA handles many data augmentations but is also computationally expensive.
By not executing data augmentation with small blend weights, the number of data augmentation executions is significantly reduced while maintaining accuracy.
The details are described in the Appendix.

\begin{table*}[tb!]
\centering
\caption{Data augmentations and magnitude parameter ranges. A-Contrast, LPFs, and HPFs denote auto-contrast, low-pass filters, and high-pass filters, respectively.}
\label{tab:augmentation_detail}
\begin{tabular}{lccc|lccc}
\toprule
Augmentation & $f^{\mathrm{act}}$ & $M$ & Inverse & Augmentation & $f^{\mathrm{act}}$ & $M$ & Inverse \\ \midrule
Rotate & tanh & 30 & Rotate & Hue & tanh & 2.0 & Hue \\
Scale & tanh & 0.3 & Scale & Equalize & - & - & Equalize \\
Saturate & sigmoid & 5.0 & Saturate & Invert & - & - & Invert \\
Contrast & sigmoid & 3.0 & Contrast & Gamma & sigmoid & 3.0 & Gamma \\
Sharpness & sigmoid & 10 & Sharpness & LPFs ($0.05,\cdots,0.95$) & - & - & HPFs \\
Brightness & tanh & 0.6 & Brightness & HPFs ($0.05,\cdots,0.95$) & - & - & LPFs \\
A-Contrast & - & - & Contrast & URIE & - & - & - \\ \bottomrule
\end{tabular}
\end{table*}

\subsection{Training and Testing}
DynTTA is trained by applying it before inference by a pretrained classification model.
When DynTTA is trained, the classification model is frozen.
The output image of DynTTA is used as input to the classification model to calculate the cross entropy loss.
Because DynTTA is differentiable, it minimizes this loss and outputs recognition-friendly images.
During testing, the DynTTA is frozen and the DynTTA output image is input into the classification model to obtain the prediction results.

\subsection{Retraining with Estimated Training-time Augmentations}
We propose a different use of DynTTA, improving accuracy by reproducing unknown distribution shifts with data augmentations and retraining the classification model with it.
For a distribution-shifted dataset, DynTTA outputs the blend weights that weight effective test-time augmentations.
We hypothesize that the distribution shift would consist of inverse operations of the highly weighted test-time augmentations.
For example, consider an unknown distribution shift dataset for which DynTTA estimates that LPF is effective. 
Since LPF removes high-frequency components, this dataset is expected to contain high-frequency components.
To make a classification model robust against high-frequency components, high-frequency components should be included in the training distribution. 
To achieve this, the inverse operation of LPF, \ie, HPF, is used to include the high-frequency components in the training dataset and retrain the model.

\Cref{algo:da_dyntta} shows how to estimate effective augmentations at training-time.
We rank the effectiveness of augmentation by the blend weights and obtain the top $k$ inverse operations of estimated augmentations.
If the inverse operation does not exist, it is omitted.
The inverse operation for each augmentation is defined in \Cref{tab:augmentation_detail}.
The estimated augmentations are included in the AugMix~\cite{augmix} augmentation space to make a classification model robust against a given unknown distribution.


\begin{algorithm}[tb!]
\DontPrintSemicolon
\KwData{A set of samples from an unknown distribution;}
\KwResult{Effective training-time augmentations for a given distribution;}
Initialize an empty list $L$;\\
\For{each estimating step}
{
Sample a mini-batch from the dataset of unknown distribution;\\
Output the blend weights and magnitude parameters by DynTTA; \\
Append the blend weights to $L$; \\
}
Average the blend weights for each augmentation in $L$;\\
Sort and rank the blend weights of the augmentations in descending order;\\
\KwRet{Inverse operations of the top $k$ augmentations}
\caption{Estimation of effective training-time augmentations}
\label{algo:da_dyntta}
\end{algorithm}

\section{Experiments}
This section describes the experimental setup, experimental results comparing DynTTA to other image enhancement methods, and detailed experiments with DynTTA.
It also shows that robustness is significantly improved by using DynTTA to estimate the effective training-time augmentations for distribution shift, which is then used to retrain the classification model.
Some experiments are described in more detail in the Appendix.

\subsection{Training and Evaluation Settings}
In this study, two settings were used for training and evaluation: blind and non-blind.
In the non-blind setting, which is used in the existing literature~\cite{lp,urie}, the corruption dataset~\cite{IN-C} is used.
This dataset consists of 19 types of corruption with five severity levels generated by the algorithm.
15 corruptions are used for training as \textit{Seen} corruptions, and four corruptions are used for testing as \textit{Unseen} corruptions.
\textit{Seen} and \textit{Unseen} corruptions consist of the same four categories (noise, blur, weather, and digital).
In this setting, the type of corruption (\eg noise) in the test set is known in advance, so this knowledge is used to generate artificial corruption (\eg Gaussian noise), which is then used as \textit{Seen} for training.
However, this setting is often impractical because test set corruption is often unknown.
Therefore, we introduce a new training and evaluation setting, the blind setting, which has no assumption on test-time distribution shifts.
In the blind setting, the distribution of the test set is unknown, so data augmentation is used such as AugMix, which improves robustness to unknown distributions.
AugMix mixes nine augmentations by default, and these augmentations do not include the corruption datasets used for testing.
In this setting, an image enhancement model trains to enhance the features needed to classify out-of-distribution data from diverse data generated by the data augmentation.
This enables the image enhancement model to increase classification accuracy for an unknown distribution, even though it does not learn the distribution of test time. 

\subsection{Experimental Setup}
Our method was evaluated on two image recognition datasets (CUB~\cite{cub} and ImageNet~\cite{imagenet}) and one domain generalization dataset (PACS~\cite{pacs}).
All experiments except ImageNet were run three times, and the average results are reported.
Loss Predictor~(LP)~($k=1$)~\cite{lp} and URIE~\cite{urie} were used as comparison methods.
Classification models were prepared in advance, and the backbone network of DynTTA and LP was prepared by fine-tuning the ResNet18~\cite{resnet} pretrained on ImageNet.
For training in the non-blind setting, corruption was randomly selected for each mini-batch to train the enhancement models.
The training corruptions consist of 15 \textit{Seen} corruptions in five levels and clean (no corruption).
For testing, four \textit{Unseen} corruptions were used.
In the blind setting, the enhancement models were trained using AugMix+DeepAugment~\cite{IN-R} for ImageNet and AugMix for the other datasets.
For testing, all 19 corruptions were used.
When evaluating on the corruptions dataset, the average accuracy was used as an evaluation metric.
In addition to evaluating the robustness of these distribution shift datasets, the clean accuracy on the standard dataset was also evaluated.
The baseline did not use the enhancement model, and the difference from the baseline is reported in the results.
In the tables showing experimental results, the best result is shown in \textbf{bold}, and the second best result is shown in \underline{underline}.

\subsection{Performance Evaluation}
\subsubsection{Classification on the CUB}
The results for classification accuracy on the CUB dataset are shown in \Cref{tab:cub}.
In the non-blind setting, URIE improves robustness but reduces the clean accuracy by a maximum of about 3.3 percentage points.
LP does not decrease the clean accuracy, but the improvement in robustness is small.
DynTTA outperforms the comparison methods in terms of both clean accuracy and robustness and significantly improves the robustness with almost no decrease in clean accuracy.
In the blind setting, similar to the non-blind setting, URIE tends to reduce the clean accuracy by a maximum of about 1.0 percentage points.
LP improves both clean accuracy and robustness slightly.
DynTTA has less degradation in clean accuracy than URIE, and it is also more robust.
In particular, DynTTA significantly improves the robustness of Mixer-B16~\cite{mlp_mixer} compared to the comparison methods.
Our experimental results show that AugMix-trained enhancement models further improve the robustness of AugMix-trained classification models.
The results indicate that our method can be used in conjunction with existing robustness improvement methods.
Furthermore, experimental results on generalizability for different classification models at training and testing, the effects of the backbone network, augmentation space, and magnitude range, and comparison to simple baselines are provided in the Appendix.

\begin{table}[tb!]
\centering
\caption{
Classification accuracy on the CUB dataset. 
The top table shows the results for the non-blind setting, and the bottom table shows the results for the blind setting. 
The numbers in parentheses indicate the differences from the baseline. 
$^\dagger$ indicates that the classification model is trained by AugMix. 
}
\label{tab:cub}
\begin{tabular}{ll|rr}
\toprule
\textbf{Non-blind} &  & & \\
Classifier & Enhancer  & Clean & Unseen \\ \midrule
ResNet50            & URIE            & 78.39\color{red}~(-3.32)  & \underline{55.93}\color{green}~(7.53) \\
                    & LP  & \textbf{81.70}\color{red}~(-0.01)  & 50.88\color{green}~(2.48) \\
                    & DynTTA          & \underline{81.58}\color{red}~(-0.13) & \textbf{58.02}\color{green}~(9.62) \\ \midrule
ResNet50$^\dagger$            & URIE            & 80.99\color{red}~(-1.56)   & \underline{63.95}\color{green}~(3.95)  \\
          & LP  & \underline{82.59}\color{green}~(0.04)  & 60.79\color{green}~(0.79) \\
                    & DynTTA          & \textbf{82.64}\color{green}~(0.09)  & \textbf{65.49}\color{green}~(5.49)  \\ \midrule
Mixer-B16          & URIE            & 86.73\color{red}~(-0.69)  & \underline{72.77}\color{green}~(3.93) \\
                    & LP  & \underline{87.39}\color{red}~(-0.03)  & 70.06\color{green}~(1.22) \\
                    & DynTTA          & \textbf{87.56}\color{green}~(0.14) & \textbf{73.89}\color{green}~(5.05)  \\ \midrule
Mixer-B16$^\dagger$           & URIE            & 86.61\color{red}~(-0.47)  & \underline{75.88}\color{green}~(2.68) \\
          & LP  & \underline{86.93}\color{green}~(0.05) & 74.20\color{green}~(1.00) \\
                    & DynTTA          & \textbf{87.26}\color{green}~(0.38) & \textbf{76.58}\color{green}~(3.38) \\ \midrule
DeiT-base          & URIE            & 83.91\color{red}~(-1.25)  & \underline{68.61}\color{green}~(1.79) \\
                    & LP  & \textbf{85.16}\color{black}~(0.00) & 67.47\color{green}~(0.65) \\
                    & DynTTA          & \underline{84.79}\color{red}~(-0.37)  & \textbf{69.80}\color{green}~(2.98) \\ \midrule
DeiT-base$^\dagger$           & URIE            & 83.88\color{red}~(-0.29)  & \underline{70.56}\color{green}~(1.50) \\
            & LP  & \underline{84.20}\color{green}~(0.03) & 69.64\color{green}~(0.58) \\
                    & DynTTA          & \textbf{84.42}\color{green}~(0.25) & \textbf{71.76}\color{green}~(2.70) \\ \bottomrule 
\multicolumn{1}{c}{}\\
\toprule
\textbf{Blind} &  & & \\
Classifier & Enhancer  & Clean & Corruption \\ \midrule
ResNet50            & URIE           & 80.68\color{red}~(-1.03)  & \underline{51.47}\color{green}~(2.89) \\
                    & LP & \textbf{81.75}\color{green}~(0.04) & 48.61\color{green}~(0.03) \\
                    & DynTTA         & \underline{81.65}\color{red}~(-0.06)  & \textbf{51.97}\color{green}~(3.39) \\ \midrule
ResNet50$^\dagger$            & URIE           & \underline{82.75}\color{green}~(0.20) & \underline{62.05}\color{green}~(0.30) \\
           & LP & 82.65\color{green}~(0.10) & 61.76\color{green}~(0.01) \\
                    & DynTTA         & \textbf{82.84}\color{green}~(0.29) & \textbf{62.39}\color{green}~(0.64) \\ \midrule
Mixer-B16          & URIE           & 87.04\color{red}~(-0.38)  & 71.27\color{green}~(0.07)    \\
                    & LP & \textbf{87.49}\color{green}~(0.07) & \underline{71.78}\color{green}~(0.58) \\
                    & DynTTA         & \underline{87.37}\color{red}~(-0.05)  & \textbf{72.74}\color{green}~(1.54) \\ \midrule
Mixer-B16$^\dagger$           & URIE           & 86.89\color{green}~(0.01) & 74.41\color{green}~(0.04) \\
         & LP & \underline{86.95}\color{green}~(0.07) & \underline{74.44}\color{green}~(0.07)  \\
                    & DynTTA         & \textbf{87.12}\color{green}~(0.24) & \textbf{75.46}\color{green}~(1.09) \\ \midrule
DeiT-base          & URIE           & 84.71\color{red}~(-0.45)  & \textbf{67.44}\color{green}~(0.96) \\
                    & LP & \textbf{85.21}\color{green}~(0.05) & 66.49\color{green}~(0.01) \\
                    & DynTTA         & \underline{85.06}\color{red}~(-0.10)  & \underline{67.37}\color{green}~(0.89) \\ \midrule
DeiT-base$^\dagger$           & URIE           & \underline{84.21}\color{green}~(0.04) & \textbf{70.38}\color{green}~(0.10)  \\
          & LP & \underline{84.21}\color{green}~(0.04)  & 70.28\color{black}~(0.00)  \\
                    & DynTTA         & \textbf{84.31}\color{green}~(0.14) & \underline{70.35}\color{green}~(0.07) \\ \bottomrule
\end{tabular}
\end{table}

\subsubsection{Classification on the ImageNet}

The results for classification accuracy on the ImageNet dataset are shown in \Cref{tab:in}.
In the non-blind setting, URIE improves robustness but reduces the clean accuracy by a maximum of about 1.6 percentage points.
LP does not decrease the ResNet50 clean accuracy, but the improvement in robustness is small.
In DeiT~\cite{deit}, LP shows a decrease in both clean accuracy and robustness.
DynTTA slightly reduces the clean accuracy, but it improves robustness more than the comparison methods.
In the blind setting, URIE and LP experiments show almost the same trend in results as in the non-blind setting.
DynTTA improves robustness over URIE without degrading the clean accuracy.

\begin{table}[tb!]
\centering
\caption{
Classification accuracy on the ImageNet dataset. 
The top table shows the results for the non-blind setting, and the bottom table shows the results for the blind setting. 
The numbers in parentheses indicate the differences from the baseline.
}
\label{tab:in}
\begin{tabular}{ll|rr}
\toprule
\textbf{Non-blind} &  & & \\
Classifier & Enhancer  & Clean & Unseen \\ \midrule
ResNet50       & URIE           & 74.56\color{red}~(-1.57) & \underline{49.05}\color{green}~(3.96) \\
               & LP & \textbf{76.12}\color{red}~(-0.01) & 46.03\color{green}~(0.94) \\
               & DynTTA         & \underline{75.89}\color{red}~(-0.24) & \textbf{50.55}\color{green}~(5.46) \\ \midrule
DeiT-base      & URIE           & 81.01\color{red}~(-0.73) & \underline{67.50}\color{green}~(2.19) \\
               & LP & \underline{81.09}\color{red}~(-0.65) & 64.22\color{red}~(-1.09)  \\
               & DynTTA         & \textbf{81.25}\color{red}~(-0.50) & \textbf{67.92}\color{green}~(2.61) \\ \bottomrule
\multicolumn{1}{c}{}\\
\toprule
\textbf{Blind} &  & & \\
Classifier & Enhancer  & Clean  & Corruption \\ \midrule
ResNet50 & URIE           & 74.71\color{red}~(-1.42) & \underline{44.19}\color{green}~(4.62)  \\
         & LP & \textbf{76.13}\color{black}~(0.00) & 40.29\color{green}~(0.72)  \\
         & DynTTA         & \underline{76.04}\color{red}~(-0.09)  & \textbf{44.87}\color{green}~(5.30)  \\ \midrule
DeiT-base & URIE           & 80.69\color{red}~(-1.06)  & \underline{62.69}\color{green}~(1.07)  \\
        & LP & \underline{80.76}\color{red}~(-0.98) & 60.28\color{red}~(-1.34)  \\
        & DynTTA         & \textbf{81.75}\color{green}~(0.01) & \textbf{62.71}\color{green}~(1.08) \\ \bottomrule
\end{tabular}
\end{table}

Furthermore, the DynTTA performance was evaluated on distribution shift datasets Stylized-ImageNet~\cite{stylized-IN}~(Stylized), ImageNet-A~\cite{IN-A}~(A), and ImageNet-R~\cite{IN-R}~(R) other than the corruption dataset in the blind setting.
The results are shown in \Cref{tab:other_dataset}.
URIE significantly decreases accuracy when the dataset is A and the classification model is DeiT, indicating that URIE is overfitting to the Corruption and R distribution.
LP barely affects the accuracy with ResNet50 and worsens the accuracy with DeiT.
While URIE and LP sometimes show decreased accuracy, DynTTA consistently improves generalization performance on these complex distribution shift environments without overfitting a particular distribution.

\begin{table}[tb!]
\centering
\caption{
Classification accuracy in the blind setting on the ImageNet dataset variants. 
The top table shows the results using ResNet50 as the classification model, and the bottom table shows the results using DeiT as the classification model. 
The numbers in parentheses indicate the differences from the baseline.
}
\label{tab:other_dataset}
\begin{tabular}{l|rrr}
\toprule
\textbf{ResNet50} &  &  &  \\ 
Enhancer & Stylized & A & R \\ \midrule
URIE & \underline{9.38}\color{green}~(2.20) & \textbf{0.59}\color{green}~(0.59) & \textbf{38.17}\color{green}~(2.00) \\
LP & 7.16\color{red}~(-0.01) & 0.01\color{green}~(0.01) & 36.18\color{green}~(0.01) \\
DynTTA & \textbf{9.80}\color{green}~(2.63) & \underline{0.16}\color{green}~(0.16) & \underline{37.48}\color{green}~(1.31) \\ \bottomrule 
\multicolumn{1}{c}{}\\
\toprule
\textbf{DeiT} &  &   &   \\
Enhancer & Stylized & A & R \\ \midrule
URIE & \underline{19.93}\color{green}~(1.91) & 24.99\color{red}~(-2.86) & \textbf{46.96}\color{green}~(1.61) \\
LP & 16.23\color{red}~(-1.80) & \underline{25.73}\color{red}~(-2.12) & 44.17\color{red}~(-1.18) \\
DynTTA & \textbf{20.12}\color{green}~(2.09) & \textbf{28.61}\color{green}~(0.76) & \underline{45.64}\color{green}~(0.29) \\ \bottomrule
\end{tabular}
\end{table}

\subsubsection{Classification on the PACS}

The effect of changing the training domain in the blind setting was evaluated, and the average accuracy is shown in \Cref{tab:pacs}.
Other experimental settings followed DomainBed~\cite{domainbed}.
Image enhancement models trained in the blind setting (trained with AugMix) are also effective for the domain generalization dataset, and DynTTA outperforms URIE in terms of average accuracy.

\begin{table*}[tb!]
\centering
\caption{
Classification accuracy of ResNet50 in the blind setting on the PACS datasets.
Each column title indicates the training domain, and the numerical values represent the average accuracy in the test domains.
The numbers in parentheses indicate the differences from the baseline.
}
\label{tab:pacs}
\begin{tabular}{l|rrrr|r}
\toprule
Enhancer & Art & Cartoon & Photo & Sketch & Avg. \\
\midrule
URIE    & 77.81\color{green}~(2.48)& 80.48\color{green}~(3.30) & \textbf{51.06}\color{green}~(5.59) & 33.60\color{green}~(8.90) & 60.74\color{green}~(5.07) \\
DynTTA  & \textbf{78.65}\color{green}~(3.32) & \textbf{80.77}\color{green}~(3.59) & 50.33\color{green}~(4.86) & \textbf{34.27}\color{green}~(9.57) & \textbf{61.01}\color{green}~(5.34) \\
\bottomrule
\end{tabular}
\end{table*}

\subsubsection{Retraining with Estimated Augmentations}
The effect of retraining with estimated augmentations by DynTTA on the PACS dataset was evaluated.
A small split of the test domain was treated as a given unknown dataset to estimate effective data augmentations.
Note that a small split was not used in the final evaluation.
For a fair comparison with an equal number of augmentations, a scenario involving original 13 augmentations (as outlined in the official AugMix code) was compared to a scenario comprising 9 default augmentations plus 4 estimated augmentations.
The accuracy in each scenario is shown in \Cref{tab:da_pacs}.
The estimated data augmentations consistently improve accuracy over the original.

\begin{table*}[tb!]
\centering
\caption{
Classification accuracy of retrained ResNet50 on the PACS datasets.
The numbers in parentheses indicate the differences from the baseline.
}
\label{tab:da_pacs}
\begin{tabular}{l|rrrr|r}
\toprule
Augmentation & Art & Cartoon & Photo & Sketch & Avg. \\
\midrule
Original                & 75.62\color{green}~(0.29) & 79.01\color{green}~(1.83) & 47.07\color{green}~(1.60) & 32.80\color{green}~(8.10)  & 58.62\color{green}~(2.95) \\
Estimated  & \textbf{78.03}\color{green}~(2.70) & \textbf{79.88}\color{green}~(2.70) & \textbf{52.84}\color{green}~(7.37) & \textbf{37.63}\color{green}~(12.93)  & \textbf{62.10}\color{green}~(6.43) \\
\bottomrule
\end{tabular}
\end{table*}

\subsection{Detailed Experiments}
This section describes experiments on the CUB dataset using ResNet50 as the classification model for an in-depth analysis of DynTTA

\subsubsection{Effects of Key Ideas}
We conducted an ablation study of our key ideas using ResNet50 on the CUB dataset and compared it to LP.
LP selects the best augmentation for a test image by learning the classification losses of 12 augmented images\footnote{Identity; \{-20, 20\} Rotate; \{0.8, 1.2\} Zoom; \{0.5, 2.0\} Saturate; Auto-contrast; \{0.2, 0.5, 2.0, 4.0\} Sharpness} as a label.
One of our key ideas, ``Blending’’, improves upon LP by using not one image but a combination of many augmented images. 
``Blending’’ is lightweight because it does not require the computation of classification losses, which has the advantage of being easily extendable in augmentation space.
Specifically, LP requires the same number of classification model inferences as the number of data augmentations used, while ``Blending’’ requires only one.
Here we define DynTTA~(BL), which has the same augmentation and magnitude parameters as LP (see footnote) and outputs only blend weights.
The effect of the ``Blending’’ was measured by using DynTTA~(BL).
In addition, to measure the effect of ``Optimization’’, we define DynTTA~(BL$+$OPT) that extends the DynTTA~(BL) and simultaneously outputs magnitude parameters.
At this time, the magnitude parameter ranges were set the same as in \Cref{tab:augmentation_detail} except for Zoom, and Identity was not used because it was included in some data augmentations (\eg, a rotation of 0 degrees).
The results are shown in \Cref{tab:2key_idea}.
DynTTA~(BL) is much more robust than LP, meaning that blending multiple images is better than choosing one best image.
In the non-blind setting, DynTTA~(BL$+$OPT) is less robust than DynTTA~(BL).
This is because the magnitude parameters used by LP and ``Blending’’ may have been tuned by the authors for their non-blind setting experiment.
On the other hand, in the blind setting, DynTTA~(BL$+$OPT) is more robust than DynTTA~(BL).
This is because ``Optimization’’ automatically finds the local optimal magnitude parameters for the blind setting using gradient descent (here we use Adam~\cite{adam}, see \Cref{sec:dyntta} and Appendix for details).
Moreover, ``Optimization’’ eliminates magnitude hyperparameters.
For example, LP has 10-magnitude hyperparameters, but these are difficult to tune in the blind setting.
``Optimization’’ eliminates these hyperparameters, making DynTTA more practical.

\begin{table*}[tb!]
\centering
\caption{
Measuring the effect of our two key ideas on the CUB dataset.
The top table shows the results for the non-blind setting, and the bottom table shows the results for the blind setting. 
The numbers in parentheses indicate the differences from the LP.
}
\label{tab:2key_idea}
\begin{tabular}{l|rrrrr}
\toprule
 & \multicolumn{2}{c}{\textbf{Non-blind}} & \multicolumn{2}{c}{\textbf{Blind}} &  \\
Enhancer &  Clean & Corruption &  Clean & Corruption &  \\
\midrule
DynTTA~(BL) & 81.64\color{red}~(-0.06) & 52.49\color{green}~(1.61) & 81.71\color{red}~(-0.04) & 51.12\color{green}~(2.51) &  \\
DynTTA~(BL$+$OPT) & 81.68\color{red}~(-0.02) & 52.15\color{green}~(1.27) & 81.73\color{red}~(-0.02) & 51.51\color{green}~(2.90) & 
\\
\bottomrule
\end{tabular}
\end{table*}

\subsubsection{Avoiding Overfitting of DNN-based Transformation}
\begin{figure*}[tb!]
\centering
\includegraphics[width=0.6\linewidth]{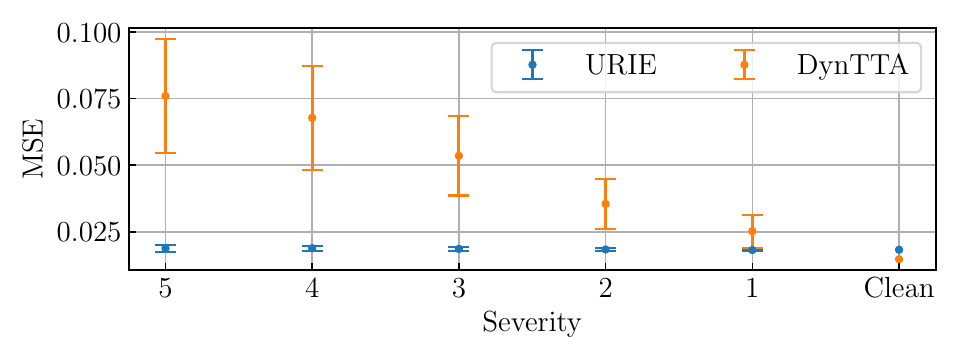}
\caption{
Measuring diverse transformations by mean and standard error of the MSE.
ResNet50 was used as the classification model in the blind setting on the CUB dataset.
}
\label{fig:mse}
\end{figure*}

When a model overfits, it tend to use the same set of transformations more (less diverse), which happens in URIE.
The mean squared error (MSE) between the raw and enhanced images was measured as the amount of image transformation to show that DynTTA reduces overfitting.
The results are shown in \Cref{fig:mse}.
URIE almost always shows the same MSE, regardless of severity, and performs almost the same set of transformations on any image, resulting in useless transformations even on clean images.
In contrast, DynTTA shows a large MSE at high severity, a small MSE at low severity, and a particularly small MSE for clean images.
Furthermore, DynTTA has a large variance of MSEs, indicating that it learns to use different transformations depending on the severity and type of corruption.
The data augmentations introduced by DynTTA work as a hint to learn diverse transformations, avoiding overfitting and enhancing images under various distributions.

\subsubsection{Backbone Network of DynTTA}
Because any neural network can be used as the backbone network of DynTTA, the effect of backbone network selection was examined.
Four models were used as backbone network: ResNet18, ResNet50~\cite{resnet}, EfficientNet-b0~\cite{efficientnet}, and MobileNetV3~\cite{mobilenetv3}.
The results are shown in the \Cref{tab:backpone}.
ResNet18 shows the highest robustness in the non-blind setting.
EfficientNet-b0 and MobileNetV3 are more robust than ResNet in the blind setting, but their clean accuracy degrades more in the non-blind setting.
ResNet50 has higher classification accuracy than ResNet18, but there is no correlation between the accuracy of the model as a classification model and the performance of the model as an enhancement model.

\begin{table}[tb!]
\centering
\caption{
Classification accuracy for each backbone network on the CUB dataset. 
The top table shows the results for the non-blind setting, and the bottom table shows the results for the blind setting. 
The numbers in parentheses indicate the differences from the baseline.
}
\label{tab:backpone}
\small
\begin{tabular}{l|rr}
\toprule
\textbf{Non-blind} & & \\ 
Backbone Network   & Clean                    & Corruption                    \\ \midrule
ResNet18           & 81.58\color{red}~(-0.13)  & 58.02\color{green}~(9.62)  \\
ResNet50           & 81.74\color{green}~(0.03) & 57.56\color{green}~(9.16)  \\
EfficientNet-b0    & 81.27\color{red}~(-0.44)  & 57.37\color{green}~(8.97)  \\
MobileNetV3        & 81.08\color{red}~(-0.63)  & 57.65\color{green}~(9.25)  \\ \bottomrule
\multicolumn{1}{c}{}\\
\toprule
\textbf{Blind} & & \\ 
Backbone Network   & Clean                    & Corruption \\ \midrule
ResNet18           & 81.65\color{red}~(-0.06)  & 51.97\color{green}~(3.39) \\
ResNet50           & 81.73\color{green}~(0.02) & 52.07\color{green}~(3.49) \\
EfficientNet-b0    & 81.73\color{green}~(0.02) & 52.95\color{green}~(4.37) \\
MobileNetV3        & 81.73\color{green}~(0.02) & 53.00\color{green}~(4.42) \\ \bottomrule
\end{tabular}
\end{table}

\subsubsection{Performance Evaluation Using a Classification Model Different from the One Used During Training}
This section discusses the generalizability of image enhancement models to classification models other than what was used for its training.
Usually, there is a one-to-one correspondence between image enhancement models and classification models.
We investigated whether an image enhancement model trained together with one classification model is also beneficial for another classification model.
The results of the image enhancement models trained with ResNet50, MLP-Mixer, and DeiT in the non-blind setting on the CUB dataset are shown in \Cref{tab:generalizability}.
URIE trained on ResNet50 significantly reduces clean accuracy and does not improve robustness.
URIE trained with MLP-Mixer or DeiT reduces clean accuracy by about 1-2 percentage points but improves robustness.
LP improves robustness with almost no reduction in clean accuracy.
DynTTA trained with ResNet50 reduces clean accuracy about 1 percentage point but improves robustness.
DynTTA trained with MLP-Mixer or DeiT improves robustness over the comparison methods with almost no reduction in clean accuracy.
Our experimental results show that when the coupled classification model is a high accuracy model such as MLP-Mixer or DeiT, the image enhancement model has high generalizability.
In particular, DynTTA improves robustness over the comparison methods with almost no reduction in clean accuracy, which shows a high generalizability.
DynTTA trained with a highly accurate classification model shows its potential to be applied to a variety of classification models.

\begin{table*}[tb!]
\centering
\caption{
Classification accuracy of the image enhancement models when the classification models differed between training and testing.
We experimented on the CUB dataset in the non-blind setting.
The numbers in parentheses indicate the differences from the baseline. 
}
\label{tab:generalizability}
\begin{tabular}{lll|rr}
\toprule
Classifier at training & Classifier at testing & Enhancer & Clean & Corruption \\ \midrule
ResNet50 & Mixer-B16 & URIE & 83.76\color{red}~(-3.66) & 68.94\color{green}~(0.10) \\
 &  & LP & 87.38\color{red}~(-0.04) & 69.91\color{green}~(1.07) \\
 &  & DynTTA & 86.55\color{red}~(-0.87) & 71.86\color{green}~(3.02) \\ \cmidrule(l){2-5} 
 & DeiT & URIE & 81.65\color{red}~(-3.51) & 64.45\color{red}~(-2.37) \\
 &  & LP & 85.13\color{red}~(-0.03) & 67.25\color{green}~(0.43) \\
 &  & DynTTA & 84.25\color{red}~(-1.01) & 67.81\color{green}~(1.09) \\ \midrule
Mixer-B16 & ResNet50 & URIE & 80.80\color{red}~(-1.01) & 52.99\color{green}~(4.59) \\
 &  & LP & 81.74\color{green}~(0.03) & 50.67\color{green}~(2.27) \\
 &  & DynTTA & 82.03\color{green}~(0.32) & 53.29\color{green}~(4.89) \\ \cmidrule(l){2-5} 
 & DeiT & URIE & 84.35\color{red}~(-0.81) & 67.61\color{green}~(0.79) \\
 &  & LP & 85.13\color{red}~(-0.03) & 67.25\color{green}~(0.43) \\
 &  & DynTTA & 85.02\color{red}~(-0.14) & 69.29\color{green}~(2.47) \\ \midrule
DeiT & ResNet50 & URIE & 79.85\color{red}~(-1.86) & 51.51\color{green}~(3.11) \\
 &  & LP & 81.72\color{green}~(0.01) & 49.95\color{green}~(1.55) \\
 &  & DynTTA & 81.60\color{red}~(-0.11) & 53.23\color{green}~(4.83) \\ \cmidrule(l){2-5} 
 & Mixer-B16 & URIE & 86.07\color{red}~(-1.35) & 71.35\color{green}~(2.51) \\
 &  & LP & 87.37\color{red}~(-0.05) & 69.55\color{green}~(0.71) \\
 &  & DynTTA & 87.14\color{red}~(-0.28) & 72.77\color{green}~(3.93) \\ \bottomrule
\end{tabular}
\end{table*}

\subsubsection{Effects of Augmentation Space}
\begin{figure}[tb!]
\centering
\includegraphics[width=1.0\linewidth]{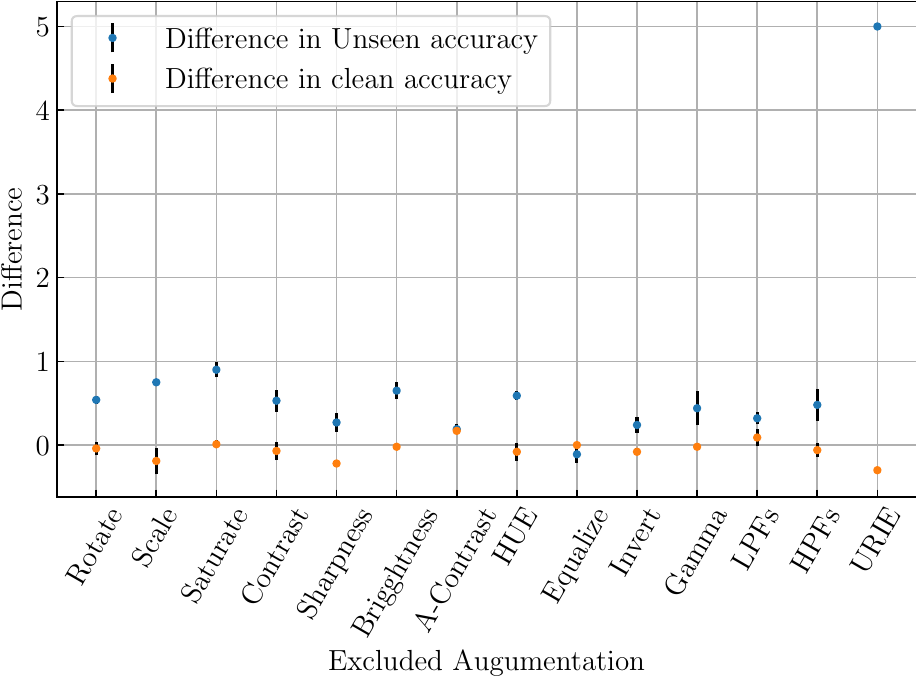}
\caption{Impact of individual augmentation on classification accuracy. The y-axis is the difference between excluding one augmentation and using all data augmentations. 
Positive values mean that they contribute to improved accuracy.
A-Contrast, LPFs, and HPFs denote auto-contrast, low-pass filters, and high-pass filters, respectively.
}
\label{fig:ab_aug}
\end{figure}

In this study, the 14 data augmentations in \Cref{tab:augmentation_detail} were used for DynTTA.
We investigated the contribution of each data augmentation to classification accuracy in the non-blind setting using ResNet50 on the CUB dataset.
\Cref{fig:ab_aug} shows the accuracy when each augmentation was excluded from DynTTA before training.
All data augmentations, except \textit{Equalize}, contribute to improving robustness with almost no reduction in clean accuracy.
In particular, URIE significantly contributed to improving robustness, but it tends to degrade clean accuracy.
DynTTA blends URIE with other data augmentations to maintain clean accuracy.
This result indicates that each data augmentation contributes differently to robustness, with some being effective and others being ineffective.
Effective data augmentations can be predefined when the type of distribution shift is known in advance, as in the non-blind setting.
For example, we estimate that \textit{Sharpness} is effective in environments where blurring often occurs.

\subsubsection{Effects of Magnitude Range}
DynTTA requires predefinition of the magnitude range $M$.
We investigated the effect of different magnitude ranges on classification accuracy.
In the experiments in the blind setting here, we also experimented with the stylized CUB dataset~\cite{stylized-IN} in addition to the corruption dataset.
The ranges in \Cref{tab:augmentation_detail} multiplied by $1.5$ and $0.5$ are referred to as ``Large’’ and ``Small’’.
The results are shown in \Cref{tab:ab_range}.
In the blind setting, ``Small’’ shows a better result for the corruption dataset and ``Large’’ for the stylized dataset. 
The clean accuracy of ``Large’’ is a little lower than the others. 
A large magnitude range is observed to be effective for large distribution shifts such as stylized datasets, but large image transformations degrade clean accuracy. 
In the non-blind setting, ``Small’’ is more robust than the standard range.
This may have been caused by ``Small’’ having a smaller search space and the optimization being more stable.

\begin{table}[tb!]
\centering
\caption{
Effect of magnitude range on classification accuracy on the CUB dataset. 
The top table shows the results for the non-blind setting, and the bottom table shows the results for the blind setting. 
The numbers in parentheses indicate the differences from the standard range DynTTA.
}
\label{tab:ab_range}
\begin{tabular}{l|rr}
\toprule
\textbf{Non-blind} &  &  \\
Range         & Clean                   & Unseen            \\ \midrule
Small & 81.58\color{black}~(0.00) & 58.41\color{green}~(0.39) \\
Large & 81.26\color{red}~(-0.32)  & 57.80\color{red}~(-0.22)  \\ \bottomrule
\end{tabular}
\begin{tabular}{l|rrr}
\multicolumn{1}{c}{}\\
\toprule
\textbf{Blind} &  & &  \\
Range         & Clean                   & Corruption & Stylized \\ \midrule
Small & 81.65\color{black}~(0.00) & 52.62\color{green}~(0.65) & 18.35\color{green}~(0.19) \\
Large & 81.61\color{red}~(-0.04) & 52.32\color{green}~(0.35) & 18.50\color{green}~(0.34) \\ \bottomrule
\end{tabular}
\end{table}

\subsubsection{Visualization of DynTTA Output}
\begin{figure*}[tb!]
\centering
\includegraphics[width=0.7\linewidth]{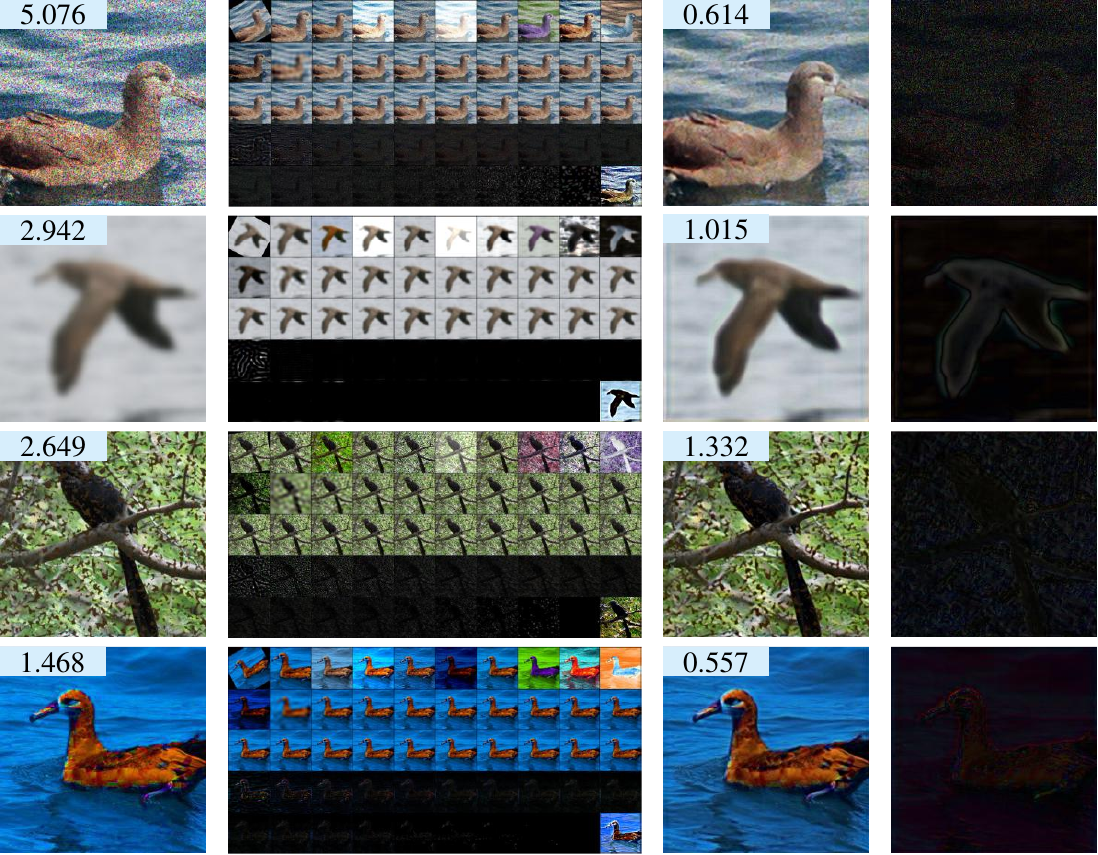}
\caption{
DynTTA output images. From top to bottom: \textit{Speckle Noise}, \textit{Gaussian Blur}, \textit{Spatter}, and \textit{Saturate}. 
The images in each row are, from left to right: input image, augmented images, output image, and difference between input and output images. 
For ease of viewing, auto-contrast is used for the high-pass filters image.
The numbers in the input and output images indicate the loss values.
}
\label{fig:out_image}
\end{figure*}

The output of DynTTA in the non-blind setting is visualized.
\Cref{fig:out_image} shows augmented images and output image by DynTTA for level-five \textit{Unseen} corruptions.
The bottom two rows of augmented images are almost black, except for the bottom right, so it is hard to see what they show, but this is because high-pass filters remove low-frequency domains that are visible to humans.
\textit{Speckle Noise} image is denoised and \textit{Gaussian Blur} image is sharpened.
The corruptions in the \textit{Spatter} and \textit{Saturate} images have not been removed, but the shape and texture of the object are enhanced.

\section{Limitations}
Although DynTTA significantly improves robustness with almost no reduction in clean accuracy, it has computational overhead.
First, to discuss network overhead, \Cref{tab:overhead} presents the number of parameters and multiply-accumulate operations (MACs).
DynTTA exhibits nearly identical numbers of parameters and MACs as URIE, yet it significantly improves performance and achieves high robustness with fewer overheads than ResNet152.
Moreover, URIE~(Large), which increases the number of layers in URIE, shows no performance improvement.
These results indicate that data augmentations integrated by DynTTA are much more effective with little overhead than increasing model size.
Next, the computational cost of performing many data augmentation runs is discussed.
The Appendix presents a technique designed to decrease the number of data augmentation runs required. 
While we acknowledge that we have not completely solved this issue, which remains a limitation of our work, we maintain that it does not affect the practical usability of DynTTA. 
As an example, utilizing the idle resources of edge devices for image enhancement effectively mitigates the concern over computational expenses~\cite{turbo}.

\begin{table}[t!]
\centering
\caption{
Classification accuracy, the number of parameters and MACs in the blind setting on the CUB dataset.
We used MobileNetV3 for the backbone of DynTTA.
}
\label{tab:overhead}
\resizebox{\linewidth}{!}{
\begin{tabular}{ll|rrrr}
\toprule
Classifier & Enhancer & Clean & Corruption & Params~(M) & MACs~(G) \\ \midrule
ResNet50 & -- & 81.71 & 48.58 & 25.56 & 4.14 \\
 & URIE & 80.68 & \underline{51.47} & 26.23 & 7.16 \\
 & URIE (Large) & 80.61 & \underline{51.47} & 37.43 & 9.96 \\
 & DynTTA & \underline{81.73} & \textbf{53.00} & 28.77 & 7.22 \\
ResNet152 & -- & \textbf{82.12} & 49.69 & 60.19 & 11.61 \\ \bottomrule
\end{tabular}
}
\end{table}

\section{Conclusion}

In this paper, we proposed DynTTA, a novel image enhancement method based on differentiable data augmentation techniques and image blending.
DynTTA uses a gradient descent algorithm to find magnitude parameters and blend weights from a huge augmentation space.
This augmentation space includes deep neural networks (DNNs) such as URIE as well as standard data augmentations.
Image enhancement with DNN and data augmentation can learn diverse transformations and thus avoid overfitting.
By applying DynTTA before inference by the pretrained classification model, DynTTA improves robustness while maintaining clean accuracy.
In addition to the existing scenario evaluated under the assumption of knowing the type of test-time distribution shift, we introduced a practical training and evaluation scenario that does not assume the type of test-time distribution shifts.
Our experimental results show that DynTTA outperforms existing methods and is effective in practical settings.
Furthermore, DynTTA estimates effective training-time augmentations for the distribution-shifted datasets and shows that retraining with estimated augmentations significantly improves accuracy.
However, the overhead during DynTTA inference has not been completely solved and is a limitation of this study.
In this study, we experimented only with the image classification task.
We will apply and evaluate DynTTA to other practical tasks such as object detection and segmentation.
Future work includes exploring the optimal backbone network architecture for DynTTA and adding differentiable data augmentation to further improve robustness.

\bibliographystyle{plain}
\bibliography{main}

\begin{IEEEbiographynophoto}
{Shohei Enomoto} received the B.S. degree from Tohoku University in 2014 and the M.S. degree from Tokyo Institute of Technology in 2016.
Since 2016, he has been engaged in researching deep learning and computer vision at NTT.
\end{IEEEbiographynophoto}

\begin{IEEEbiographynophoto}
{Monikka Roslianna Busto} is a Researcher at NTT Software Innovation Center. She graduated from the Electrical and Electronics Engineering Institute, College of Engineering, the University of the Philippines in 2017, and received a master's degree from the Department of Information and Communications Engineering, Tokyo Institute of Technology in 2021. 

She joined Nippon Telegraph and Telephone Corporation in the same year. Her research interests include computer vision, collaborative intelligence for edge computing, remote sensing image analysis and multi-modal AI.
\end{IEEEbiographynophoto}

\begin{IEEEbiographynophoto}
{Takeharu Eda} received his B.S. in Mathematics from Kyoto University in 2001 and his M.S. in Engineering from the Nara Institute of Science and Technology in 2003. He joined NTT Laboratories in 2003, focusing on research in various aspects of machine learning and systems.

In 2011, he transitioned to NTT Communications, where he launched a web hosting service utilizing CloudStack/OpenStack-based infrastructure and migration tools. He managed international development teams with members from the US, Germany, and Japan.

In 2015, he moved to the NTT Software Innovation Center, developing a scalable surveillance video system using deep learning-based computer vision techniques. Since 2022, he has been involved in the research and development of a space computing project with SpaceCompass.

Eda is a member of the Information Processing Society of Japan (IPSJ), the Association for Computing Machinery (ACM), and IEEE.
\end{IEEEbiographynophoto}

\EOD
\end{document}


\appendix
\history{Date of publication xxxx 00, 0000, date of current version xxxx 00, 0000.}
\doi{10.1109/ACCESS.2024.0429000}

\title{Supplementary Material for Dynamic Test-Time Augmentation via Differentiable Functions}

\author{\uppercase{Shohei Enomoto}\authorrefmark{1}, 
\uppercase{Monikka Roslianna Busto}\authorrefmark{1}, and \uppercase{Takeharu Eda}\authorrefmark{1},
\IEEEmembership{Member, IEEE}}

\address[1]{NTT, Musashino, Tokyo 180-8585, Japan}

\tfootnote{}

\markboth
{S. Enomoto \headeretal: Supplementary Material for Dynamic Test-Time Augmentation via Differentiable Functions}
{S. Enomoto  \headeretal: Supplementary Material for Dynamic Test-Time Augmentation via Differentiable Functions}

\corresp{Corresponding author: Shohei Enomoto (e-mail: alesana882@gmail.com).}

\maketitle

\setcounter{table}{12}
\setcounter{figure}{4}
\setcounter{equation}{3}

In this Supplementary Material, we provide additional analysis for the DynTTA. 

\section{Related Work}
\subsection{Applications of Image Enhancement}
We present applications using image enhancement.
Image enhancement has the benefit of not requiring the classification model to be retrained.
By taking advantage of this benefit, use cases have been proposed to improve the performance of models in the cloud, where only APIs are provided~\cite{salman2020denoised,blackvip}.
In addition, model retraining involves the risk of catastrophic forgetting~\cite{parisi2019continual,ewc,ebrahimi2020adversarial}, but image enhancement avoids this, making image enhancement more useful than model updating~\cite{bts,prabhudesai2024test,decorate}.

\section{Computational Cost Reduction for DynTTA}
DynTTA is more computationally expensive than Loss Predictor because it performs many data augmentations during inference.
To reduce the inference cost of DynTTA, we tried to avoid executing data augmentations with small blend weights.
With this method, we achieved a 72\% reduction in the number of data augmentation executions while maintaining the accuracy of DynTTA.
We describe the details here.

First, we set a threshold.
If all the blend weights of a data augmentation in a mini-batch are below the threshold, we do not execute that data augmentation.
Since the blend weights should add up to 1, the blend weights for the data augmentations not executed at this time are added to the blend weights of the next data augmentation that is executed.
ResNet50 was used in the non-blind setting on the CUB corruption dataset at severity level 5 to measure the average number of data augmentation executions per inference and accuracy of DynTTA.

Results are shown in \Cref{tab:num_aug}.
When the threshold is 0, DynTTA executes all 50 data augmentations.
When the threshold is less than 0.01, DynTTA reduces the average number of data augmentation executions further improving accuracy.
We suspect this to be because we were able to prevent the execution of data augmentations that should not have been executed, which were output with blend weights near zero.
When the threshold is 0.05, DynTTA reduces the number of data augmentation executions by about 72\% while maintaining accuracy.
When the threshold is greater than 0.1, DynTTA reduces the number of data augmentation executions, but it also reduces the accuracy.

\begin{table}[tb!]
\centering
\caption{Classification accuracy and average number of data augmentation executions per inference when varying the threshold. The numbers in parentheses indicate the differences from the original DynTTA (threshold is 0).}
\label{tab:num_aug}
\small
\begin{tabular}{cccc}
\hline
 & Number of & & \\
Threshold & Executions & Clean & Corruption \\ \hline
0.005     & 34.76\color{green}~(-15.24)  & 81.58\color{green}~(0.05) & 36.61\color{green}~(0.71) \\
0.01      & 28.71\color{green}~(-21.29)  & 81.65\color{green}~(0.12) & 36.68\color{green}~(0.78) \\
0.02      & 21.85\color{green}~(-28.15)  & 81.36\color{red}~(-0.17) & 35.81\color{red}~(-0.09) \\
0.05      & 13.94\color{green}~(-36.06)  & 81.15\color{red}~(-0.38) & 35.92\color{green}~(0.02) \\
0.1       & 9.29\color{green}~(-40.71)  & 80.95\color{red}~(-0.58) & 34.58\color{red}~(-1.32) \\
0.2       & 3.88\color{green}~(-46.12)  & 80.88\color{red}~(-0.65) & 19.89\color{red}~(-16.01) \\ \hline
\end{tabular}
\end{table}

\section{Details of Dataset and Training}
\subsection{Classification on the CUB}
CUB is an image dataset of 200 class bird species.
It consists of 5994 training images and 5794 test images.
We used ResNet50~\cite{resnet}, Mixer-B16~\cite{mlp_mixer} and DeiT-base~\cite{deit} as the classification model. 
For pretrained ResNet we used the weight trained on IMAGENET1K\_V1 from torchvision.models~\cite{torchvision}.
For pretrained Mixer and Deit, we downloaded the weight from official github.
For optimization, we employ Adam~\cite{adam} with learning rate $0.001$ for URIE, $0.001$ for DynTTA and Loss Predictor, and decay the learning rate every 10 epochs by 2.
These models are trained in $60$ epochs.

\subsection{Classification on the ImageNet}
ImageNet is the most widely used 1000-class image recognition dataset.
It is composed of 1.3 million training images and 50,000 test images.
We used ResNet50 and DeiT as the classification model.
For optimization, we employ Adam with learning rate $0.001$ for URIE, $0.0001$ for DynTTA and Loss Predictor, and decay the learning rate every 8 epochs by 10.
These models are trained in $30$ epochs.

\subsection{Classification on the Domain Generalization Datasets}
\begin{figure}[tb!]
\centering
\includegraphics[width=1.0\linewidth]{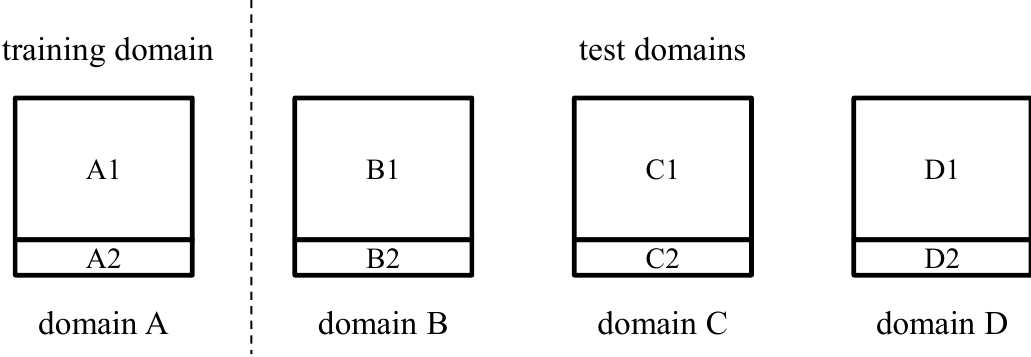}
\caption{
Dataset split for training, validation, and testing.
We split each domain into an 8:2 ratio of training or testing and validation.
For training, we use the big split of the training domain for training and the small split for validation.
For testing, we use the big split of the test domains for testing.
We use the small split of the test domains during training-time augmentation estimation with DynTTA.
}
\label{fig:split}
\end{figure}

PACS consists of four domains (art, cartoons, photos, sketches), 9,991 images and 7 classes.
VLCS~\cite{vlcs} consists of four photographic domains (Caltech101, LabelMe, SUN09, VOC2007), 10,729 images and 5 classes.
OfficeHome~\cite{officehome} consists of four domains (art, clipart, product, real), 15,588 images and 65 classes.
TerraIncognita~\cite{terraincognita} consists of four camera trap domains (L100, L38, L43, L46), 24,788 images and 10 classes.
We used ResNet50 as the classification model.
For optimization, we employ SGD with learning rate $0.0005$ and batch size is $96$.
Since we used DomainBed, all other settings follow DomainBed.
We show in the \Cref{fig:split} the details of splitting the dataset for training, validation, and testing.

\section{Effects of Optimization and Blending}
We show the results of the simple test-time augmentation using random magnitude parameters and random blend weights in \Cref{tab:random_m_w}.
We used the same data augmentation as DynTTA.
All test-time augmentations using random magnitude are less accurate, except for A-Contrast.
Random blending is also less accurate than without test-time augmentation.
These results indicate the validity of the magnitudes and weights estimated by DynTTA.

\begin{table}[tb!]
\centering
\caption{
Comparison with random magnitude parameters and random blend weights on the CUB dataset using ResNet50.
Each augmentation uses random magnitude, except augmentations that do not have magnitude.
Random blending blends each augmentation of random magnitude with a random weight.
}
\label{tab:random_m_w}
\begin{tabular}{l|cc}
\toprule
Test-time Augmentation & Clean & Corruption \\ \midrule
-- & 81.71 & 48.40 \\ \midrule
Rotate & 77.42 & 41.82 \\
Scale & 78.54 & 45.52 \\
Saturate & 59.95 & 32.81 \\
Contrast & 59.35 & 30.04 \\
Sharpness & 73.42 & 37.90 \\
Brightness & 64.29 & 32.27 \\
A-Contrast & 81.99 & 51.56 \\
Hue & 40.92 & 20.70 \\
Equalize & 50.31 & 27.35 \\
Invert & 10.92 & 4.49 \\
Gamma & 62.92 & 39.26 \\
LPFs & 69.19 & 40.62 \\
HPFs & 2.24 & 0.62 \\ \midrule
Random blending & 75.00 & 42.02 \\ \bottomrule
\end{tabular}
\end{table}

\section{Comparison with Simple Test-Time Augmentation}
The Loss Predictor paper made a comparison with some simple test-time augmentation (center crop, horizontal flip, random augmentation selection), and the results are that the Loss Predictor is better. 
Therefore, DynTTA should also be better than simple test-time augmentation. 
On the other hand, training time augmentation, which randomly combines (blends) multiple images to improve robustness (\eg, AugMix~\cite{augmix}, MixUp~\cite{mixup}, has been proposed.
Random blending of multiple images during training has the effect of improving robustness, however, it is not clear whether it has the same effect during testing.
To verify the effect of randomly blending images during testing, we experimented with test-time AugMix using ResNet50 in the blind setting on the CUB dataset.
The resulting clean accuracy is 67.64\% and corruption accuracy is 36.12\%, which is significantly worse than the baseline (without image enhancement model).
This result indicates that simple randomly blending is ineffective and that blending trained by DynTTA is effective in improving robustness.

\section{Comparison with Training Robust Model Algorithms}
We compare DynTTA with algorithms for training robust classification models.
AguMix, MixUp, and CutMix are data augmentations that improve data diversity by mixing multiple images.
MixProp, Fast AdvProp, and AdvProp are disentangled learning methods that improve robustness by training out-of-distribution data through auxiliary batch normalization layers.
We show the results in \Cref{tab:vs_train}.
Note here that we used 15 corruptions in our evaluation.
DynTTA can be used in conjunction with these algorithms to further improve robustness.

\begin{table}[tb!]
\centering
\caption{
Comparison with robust training algorithms on the CUB dataset.
We trained DynTTA in the blind setting.
}
\label{tab:vs_train}
\begin{tabular}{l|cc}
\toprule
Algorithm & Clean & 15 Corruption \\ \midrule
Vanilla & 81.71 & 48.40 \\
AugMix & 82.55 & \underline{59.97} \\
MixUp & \underline{83.04} & 58.52 \\
CutMix & 81.65 & 48.11 \\
MixProp & \textbf{83.77} & 56.80 \\
Fast AdvProp & 82.90 & 51.22 \\
AdvProp & 81.45 & 53.44 \\
Vanilla+DynTTA & 81.65 & 51.38 \\
AugMix+DynTTA & 82.84 & \textbf{60.59} \\ \bottomrule
\end{tabular}
\end{table}

\section{Classification on the Other Domain Generalization Datasets}
We show the experimental results on the VLCS dataset, the OfficeHome dataset and the TerraIncognita dataset in \Cref{tab:domainbed}.
DynTTA outperforms URIE on all datasets except VLCS.

\begin{table*}[tb!]
\centering
\caption{
Classification accuracy of ResNet50 in the blind setting on the domain generalization datasets.
}
\label{tab:domainbed}
\begin{tabular}{l|cccc|c}
\toprule
Enhancer        & VLCS            & PACS             & OfficeHome       & TerraIncognita   & Average              \\
\midrule
URIE & \textbf{63.94}\color{green}~(2.40)      & 60.74\color{green}~(5.07) & 55.69\color{green}~(0.41) & 33.75\color{green}~(1.27) & 53.53\color{green}~(2.29)  \\
DynTTA & 63.91\color{green}~(2.37) & \textbf{61.01}\color{green}~(5.34) & \textbf{55.77}\color{green}~(0.49) & \textbf{34.87}\color{green}~(1.39) & \textbf{53.89}\color{green}~(2.65)  \\
\bottomrule
\end{tabular}
\end{table*}

\section{Retraining with Estimated Augmentation on the CUB}
\begin{figure}[tb!]
\centering
\includegraphics[width=1.0\linewidth]{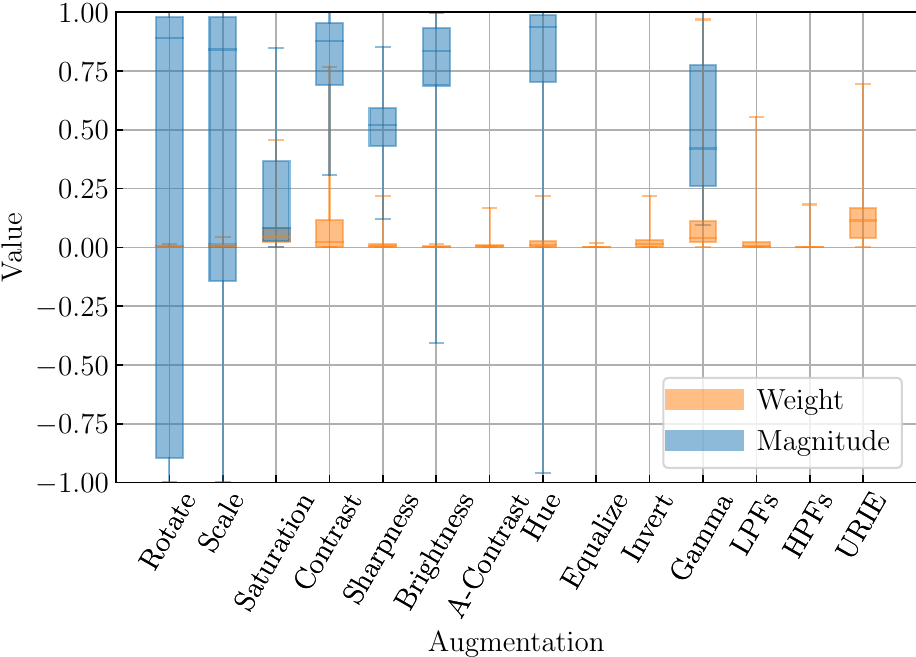}
\caption{Boxplot of magnitude parameters and blend weights for validation set. 
}
\label{fig:valset_box_m_w}
\end{figure}

In the blind setting, we split the level five four corruption (speckle noise, Gaussian blur, spatter, and saturation) test set into a validation set and a test set in a 2:8 ratio.
\Cref{fig:valset_box_m_w} shows a visualization of the magnitude parameters and the blend weights, which are the outputs of DynTTA for the validation set.
DynTTA outputs high blend weights of low-pass filters, which are effective for high-frequency corruptions such as \textit{Speckle Noise} and \textit{Spatter}, and of saturation, which is effective for \textit{Saturate}.
The maximum weights of low-pass filters, gamma, contrast, and saturation are high for validation set.
The magnitude parameters of rotate and scale have high variance.
This is because when the blend weight is close to zero, any value of the magnitude parameter has almost no effect on the output image.

Next, from the test-time augmentation that was effective for the validation set, we estimate the training-time augmentation that is effective for the target corruption.
Based on the results in \Cref{fig:valset_box_m_w}, we focus on the maximum value of the blend weights rather than the magnitude parameter.
We choose the four data augmentations used to retrain the classification model: high-pass filters, gamma, contrast, and saturation, which are the inverse operations of four data augmentations that were effective for validation set.
We experiment by varying the data augmentation space used by the AugMix.
The results of the test set are shown in \Cref{tab:interpretation}.
``Normal’’ is the normal AugMix and uses nine data augmentations \footnote{Auto-Contrast, Equalize, Posterize, Rotate, Solarize, ShearX, ShearY, TranslateX, TranslateY}.
``All’’ is the normal nine data augmentations plus saturation, contrast, brightness and sharpness, referenced from the official AugMix implementation and ``Estimated’’ is the normal nine data augmentations plus saturation, contrast, high-pass filters and gamma.
The classification model trained with ``Estimated’’ achieved the best accuracy for all corruptions except \textit{Saturate}.
On average, it was approximately 10 points more accurate than ``Normal’’, and, in particular, approximately 20 points more accurate than ``Normal’’ for \textit{Speckle Noise}.

\begin{table*}[tb!]
\centering
\caption{
Classification accuracy on level five test set when varying the augmentation space used by the AugMix. Average indicates the average accuracy of the four corruptions.
}
\label{tab:interpretation}
\begin{tabular}{l|ccccc}
\toprule
Augmentation & Speckle Noise & Gaussian Blur & Spatter & Saturate & Average \\
\midrule
Normal & \underline{50.27} & 30.49 & 35.33 & 36.75 & 38.21 \\
All & 49.90 & \underline{33.18} & \underline{35.61} & \textbf{44.97} & \underline{40.92} \\
Estimated & \textbf{70.25} & \textbf{37.22} & \textbf{43.36} & \underline{42.74} & \textbf{48.39}
\\
\bottomrule
\end{tabular}
\end{table*}

\section{URIE Size and Performance}
We show the relationship between the size and performance of URIE in \Cref{tab:urie_scal}
URIE (small) means one SKDown and one SKUp removed.
URIE (Midium) means one additional SKDown and one additional SKUp.
URIE (Large) means two additional SKDown and two additional SKUp.
The performance of URIE does not change whether the size is increased or decreased.
Changing the size of the URIE does not solve the problem of overfitting.

\begin{table}[tb!]
\centering
\caption{
Classification accuracy, the number of parameters and MACs in the blind setting on the CUB dataset.
We use ResNet50 as the classification model.
}
\label{tab:urie_scal}
\begin{tabular}{l|cccc}
\toprule
 Enhancer & Clean & Corruption & Params~(M) & MACs~(G) \\ \midrule
  URIE (Small) & 80.37 & 51.15 & 25.67 & 5.75 \\
  URIE & \textbf{80.68} & \textbf{51.47} & 26.23 & 7.16 \\
  URIE (Medium) & \underline{80.61} & \underline{51.21} & 28.47 & 8.56 \\
  URIE (Large) & \underline{80.61} & \textbf{51.47} & 37.43 & 9.96 \\\bottomrule
\end{tabular}
\end{table}

\section{Ensemble of Multiple Data Augmentations}
Loss Predictor improves accuracy by ensembling $k$ candidate augmentations and  further improves accuracy by ensembling with hflip.
In \Cref{tab:detailed_result} we show the accuracy of Loss Predictor ($k=2$) in the non-blind setting, and also in combination with hflip.

The increase in clean accuracy and robustness with $k=2$ is $+0.44$ pt and $+0.09$ pt, respectively, and $k=1$ was used in all main experiments due to the significantly lower robustness of Loss Predictor ($k=2$) compared to URIE and DynTTA.
Loss Predictor also improved clean accuracy and robustness by $+0.84$ pt and $+0.69$ pt, respectively, when combined with hflip.
However, hflip is also applicable to DynTTA, which improved by $+0.74$ pt and $+0.70$ pt, respectively.

\begin{table}[tb!]
\centering
\caption{Additional experimental results for ResNet50 on the CUB dataset in the non-blind setting. The numbers in parentheses indicate the difference from the normal enhancement. LP indicates Loss Predictor.}
\label{tab:detailed_result}
\begin{tabular}{l|cc}
\hline
Enhancer          & Clean   & Corruption  \\ \hline
LP($k=2$)              & 82.14\color{green}~(0.44) & 50.97\color{green}~(0.09) \\
LP($k=2$)$+$hflip        & 82.54\color{green}~(0.84) & 51.57\color{green}~(0.69) \\\hline
DynTTA$+$hflip         & 82.32\color{green}~(0.74) & 58.72\color{green}~(0.70) \\ \hline
\end{tabular}
\end{table}

\section{Potential integration of DynTTA with other machine learning frameworks}
Even if it is a machine learning algorithm other than a neural network, DynTTA can be applied to any algorithm that is trained by gradient descent with a loss function.
Furthermore, if a differentiable data augmentation can be defined for modalities other than image (e.g., language), DynTTA can be applied.

\section{Visual Comparison}
\begin{figure*}[tb!]
\centering
\resizebox{\linewidth}{!}{
\begin{tabular}{ccccc}
Input & URIE & Input-URIE & DynTTA & Input-DynTTA \\
\includegraphics[width=0.3\linewidth]{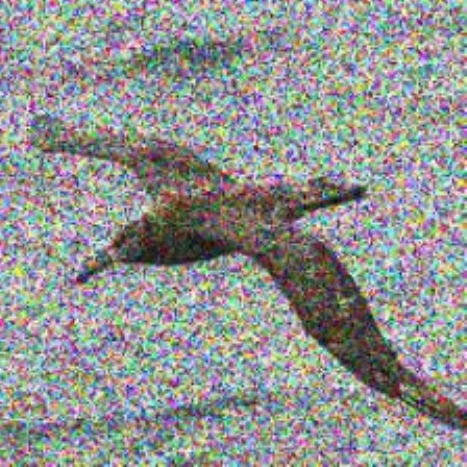} &
\includegraphics[width=0.3\linewidth]{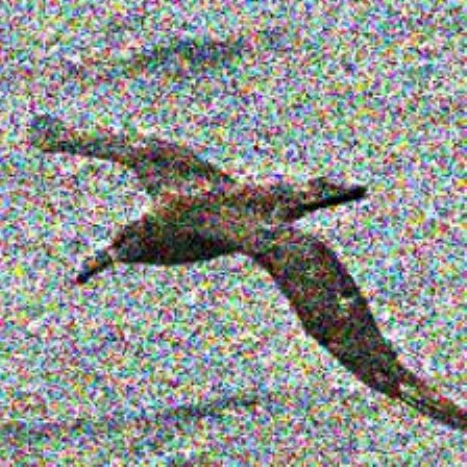}  &
\includegraphics[width=0.3\linewidth]{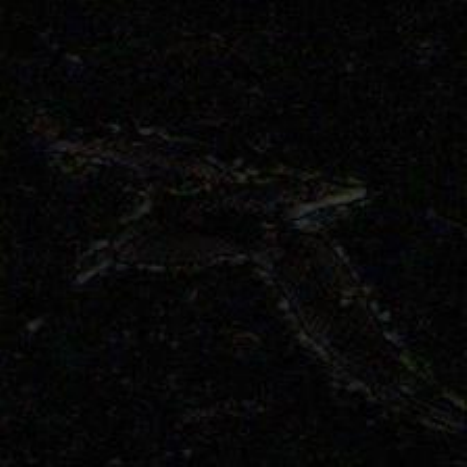}  &
\includegraphics[width=0.3\linewidth]{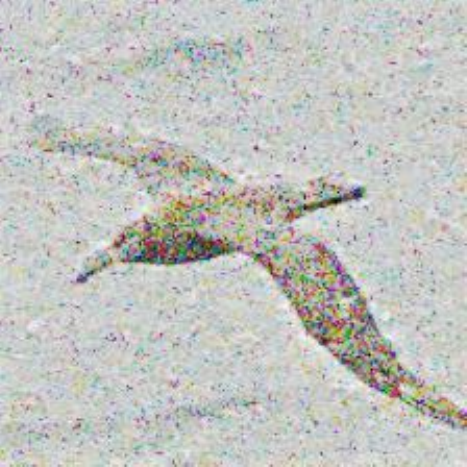}  &
\includegraphics[width=0.3\linewidth]{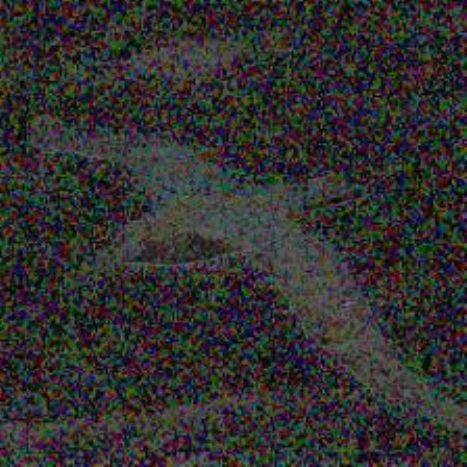}  
\\
\includegraphics[width=0.3\linewidth]{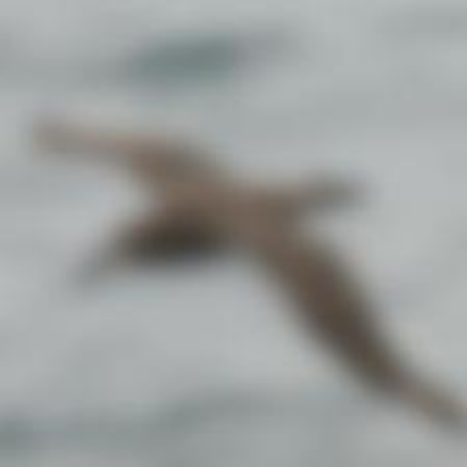} &
\includegraphics[width=0.3\linewidth]{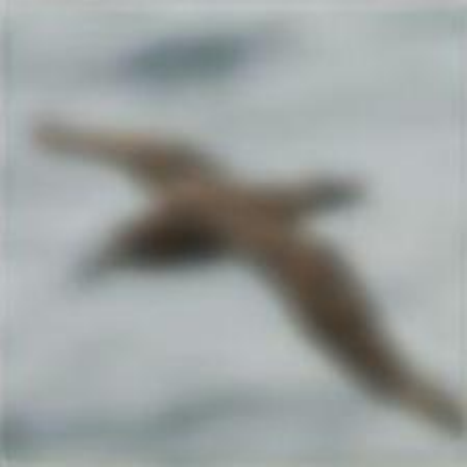}  &
\includegraphics[width=0.3\linewidth]{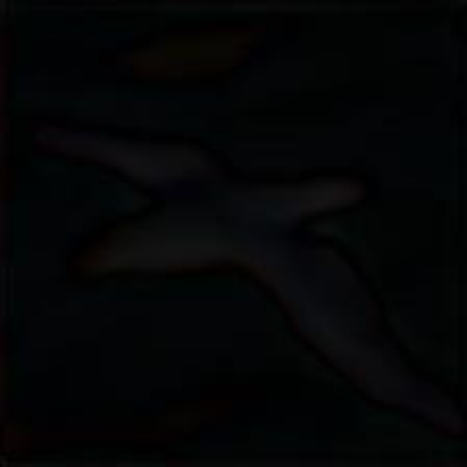}  &
\includegraphics[width=0.3\linewidth]{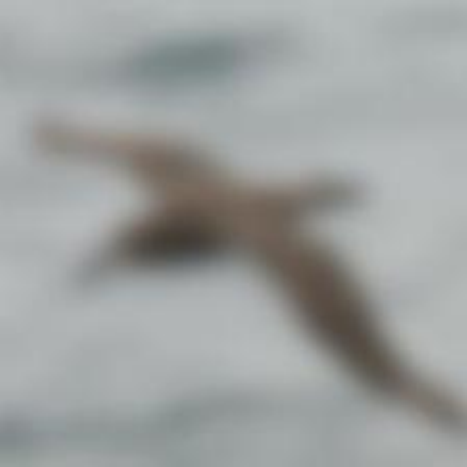}  &
\includegraphics[width=0.3\linewidth]{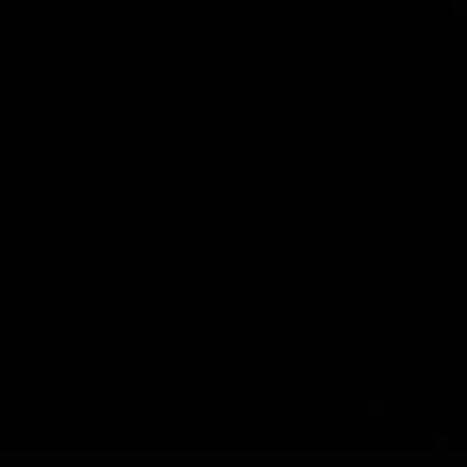}  
\\
\includegraphics[width=0.3\linewidth]{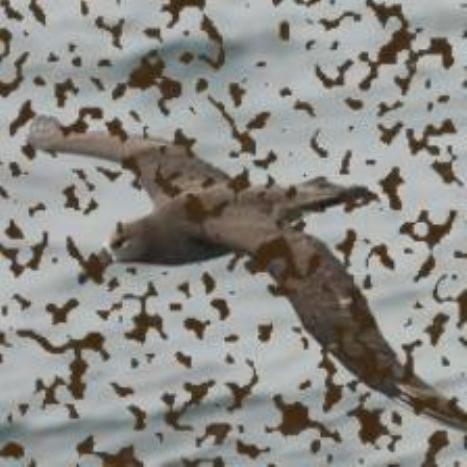} &
\includegraphics[width=0.3\linewidth]{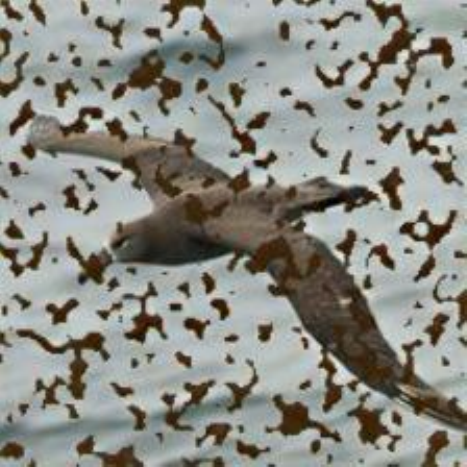}  &
\includegraphics[width=0.3\linewidth]{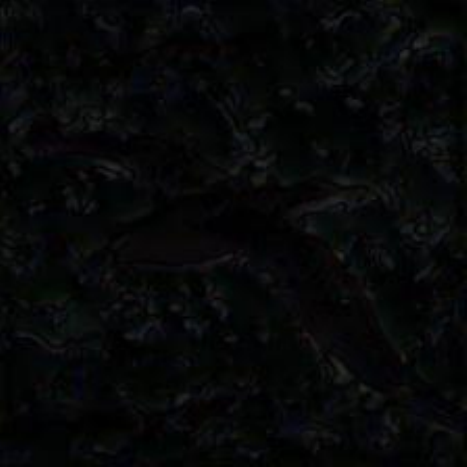}  &
\includegraphics[width=0.3\linewidth]{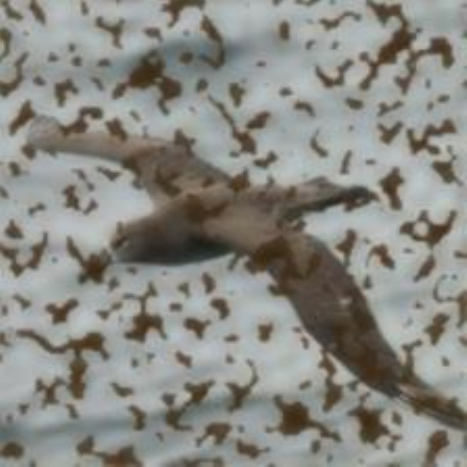}  &
\includegraphics[width=0.3\linewidth]{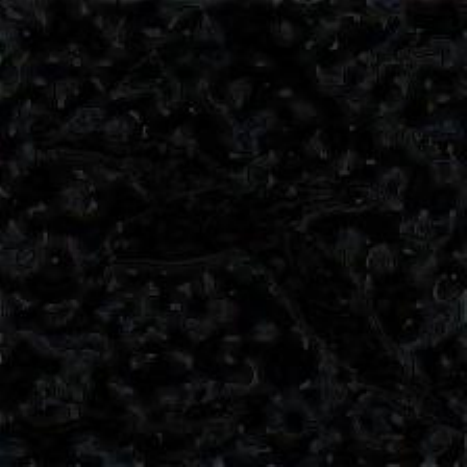}  
\\
\includegraphics[width=0.3\linewidth]{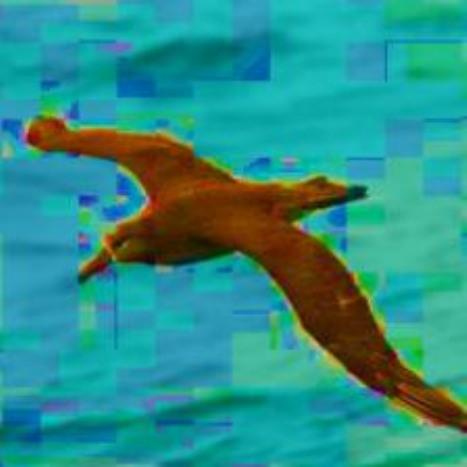} &
\includegraphics[width=0.3\linewidth]{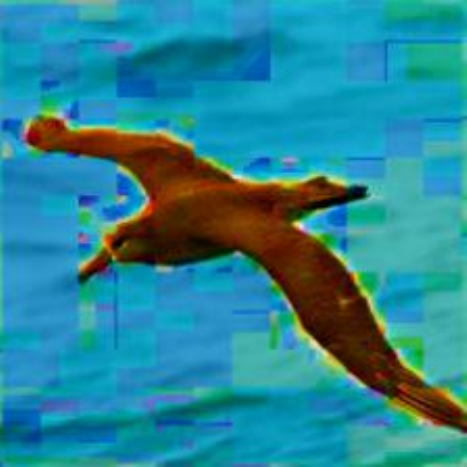}  &
\includegraphics[width=0.3\linewidth]{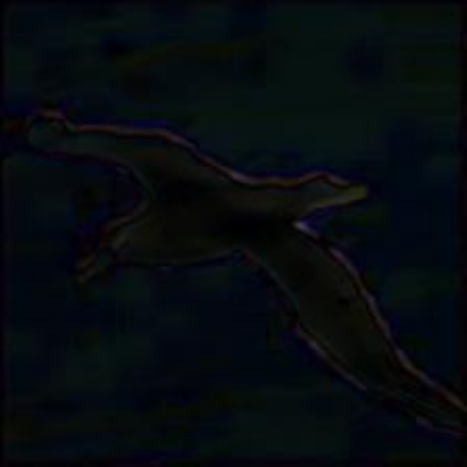}  &
\includegraphics[width=0.3\linewidth]{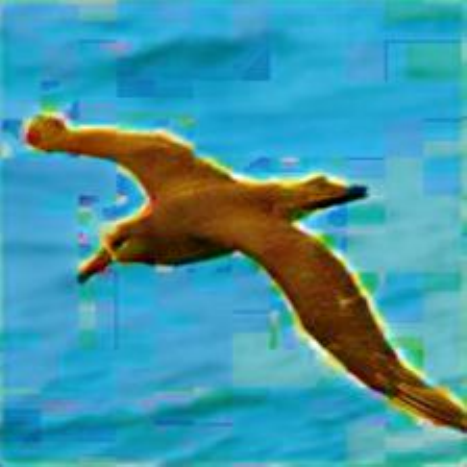}  &
\includegraphics[width=0.3\linewidth]{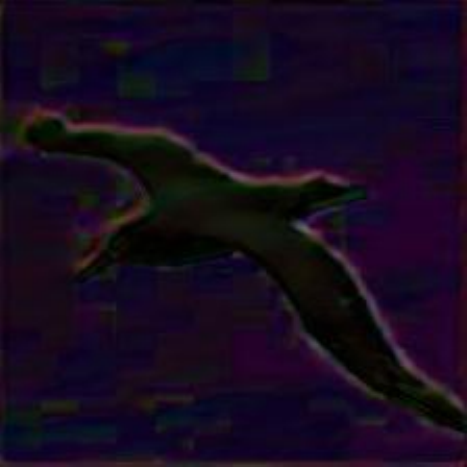}  
\\
\end{tabular}
}
\caption{
From top to bottom: speckle noise, Gaussian blur, spatter, and saturation, all at a severity level of five.
We trained the enhancer in the blind setting.
}
\label{fig:visualize}
\end{figure*}

We show in \Cref{fig:visualize} the corrupted image, the enhanced image, and the difference between the enhanced image and the corrupted image.
URIE transforms any corruption in such a way that the object shape is enhanced.
DynTTA performs noise-removing transformations for noise corruption and color transformations for saturation corruption.
DynTTA performs diverse transformations depending on the type of corruption.

\section{Details of the CUB Dataset Experiment}
\begin{figure*}[tb!]
\centering
\includegraphics[width=1.0\linewidth]{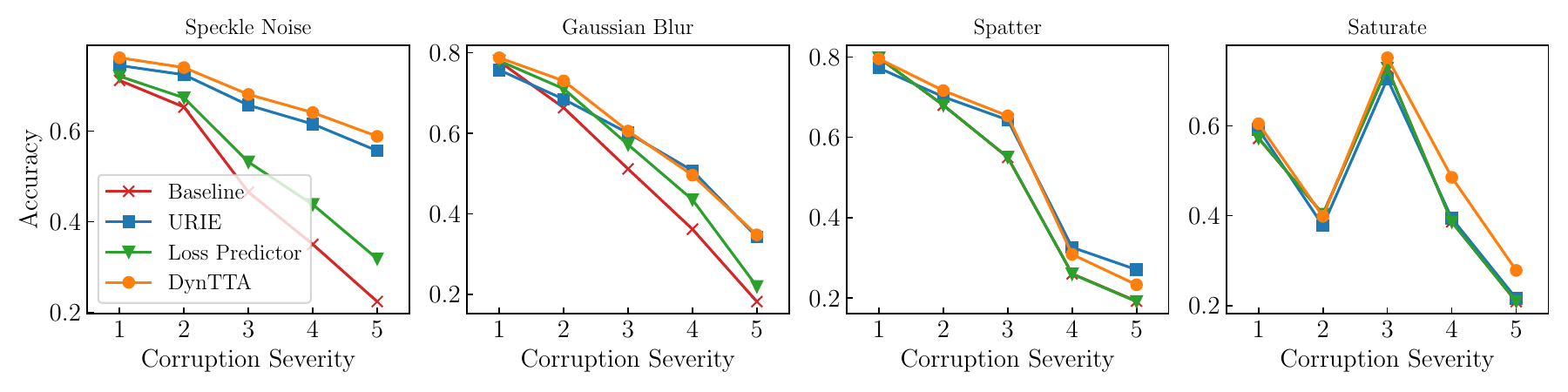}
\caption{Classification accuracy of ResNet50 for each corruption and severity on the CUB dataset.}
\label{fig:nonblind_r50}
\centering
\includegraphics[width=1.0\linewidth]{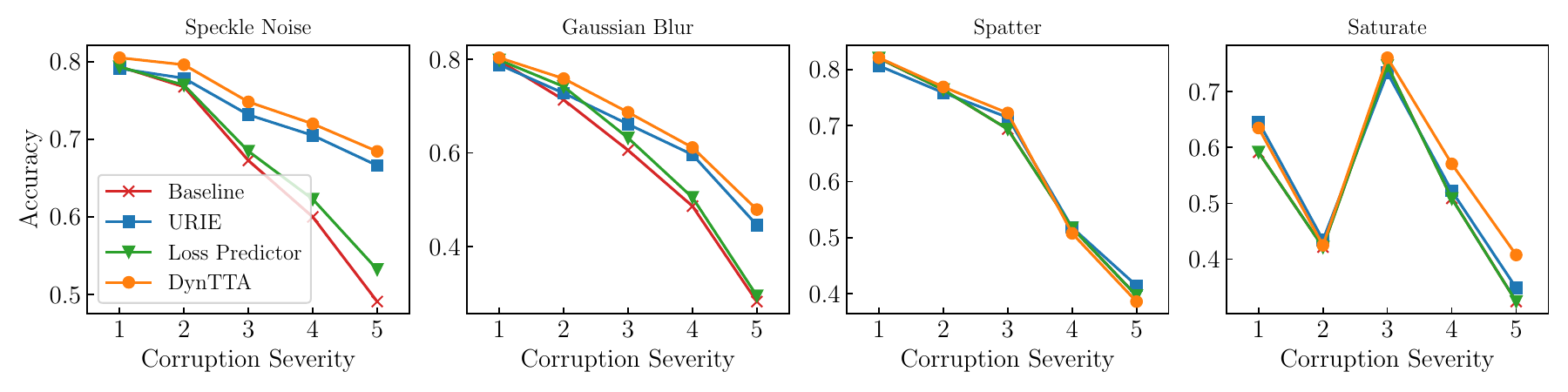}
\caption{Classification accuracy of ResNet50 w/AugMix for each corruption and severity on the CUB dataset.}
\label{fig:nonblind_r50_augmix}

\centering
\includegraphics[width=1.0\linewidth]{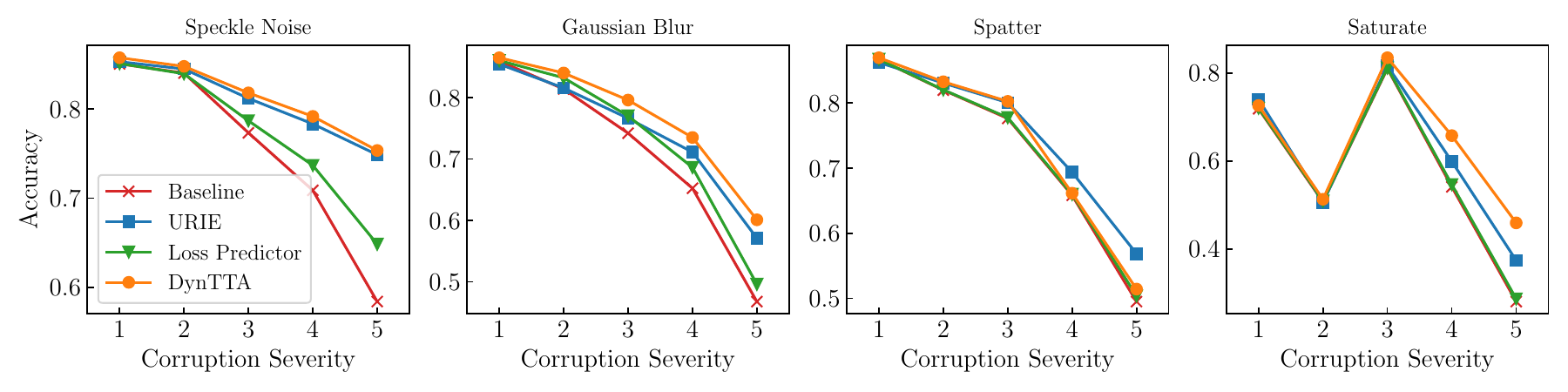}
\caption{Classification accuracy of Mixer-B16 for each corruption and severity on the CUB dataset.}
\label{fig:nonblind_mixer}
\centering
\includegraphics[width=1.0\linewidth]{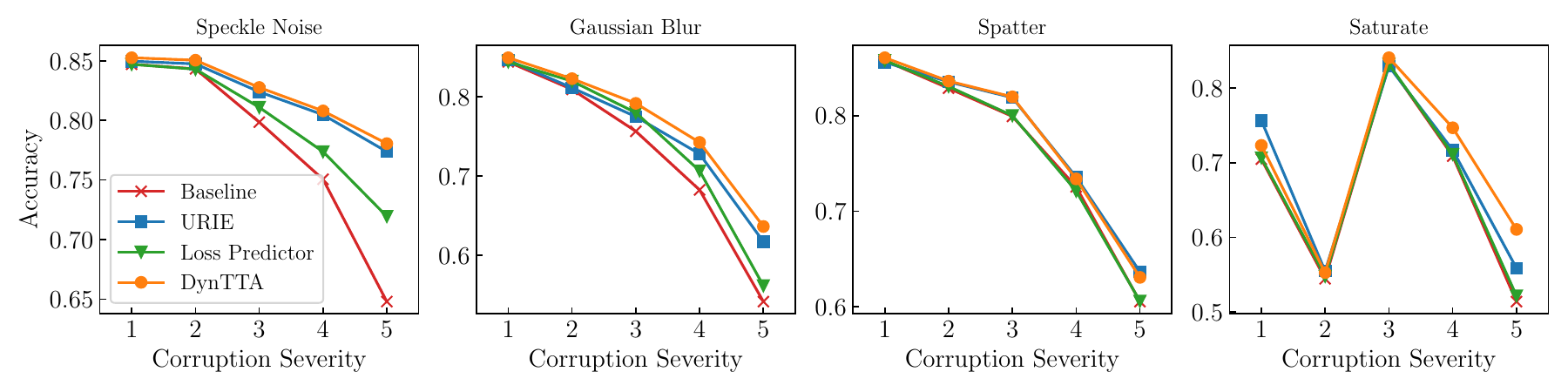}
\caption{Classification accuracy of Mixer-B16 w/AugMix for each corruption and severity on the CUB dataset.}
\label{fig:nonblind_mixer_augmix}
\end{figure*}

\begin{figure*}[tb!]
\centering
\includegraphics[width=1.0\linewidth]{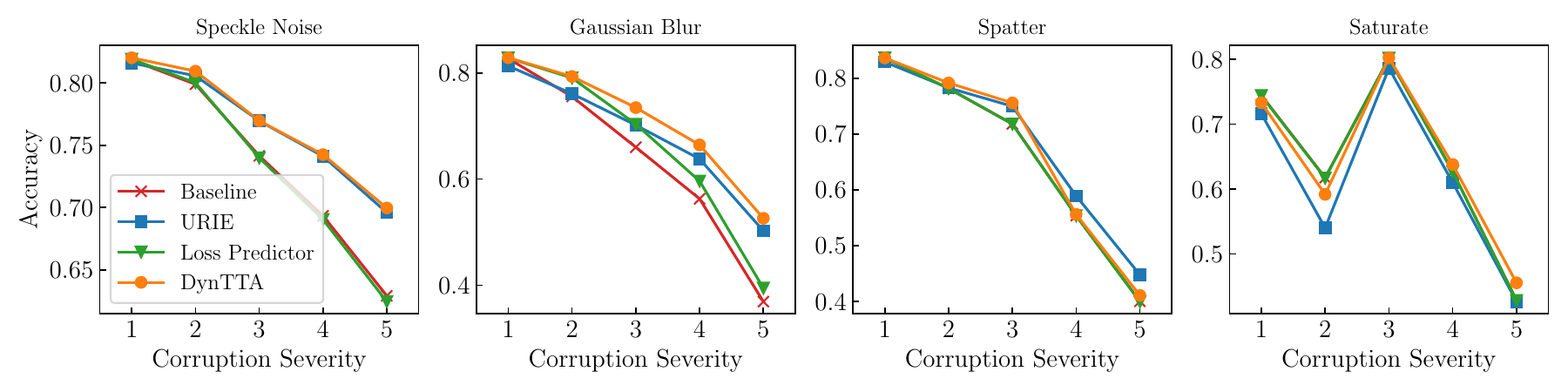}
\caption{Classification accuracy of DeiT for each corruption and severity on the CUB dataset.}
\label{fig:nonblind_deit}
\centering
\includegraphics[width=1.0\linewidth]{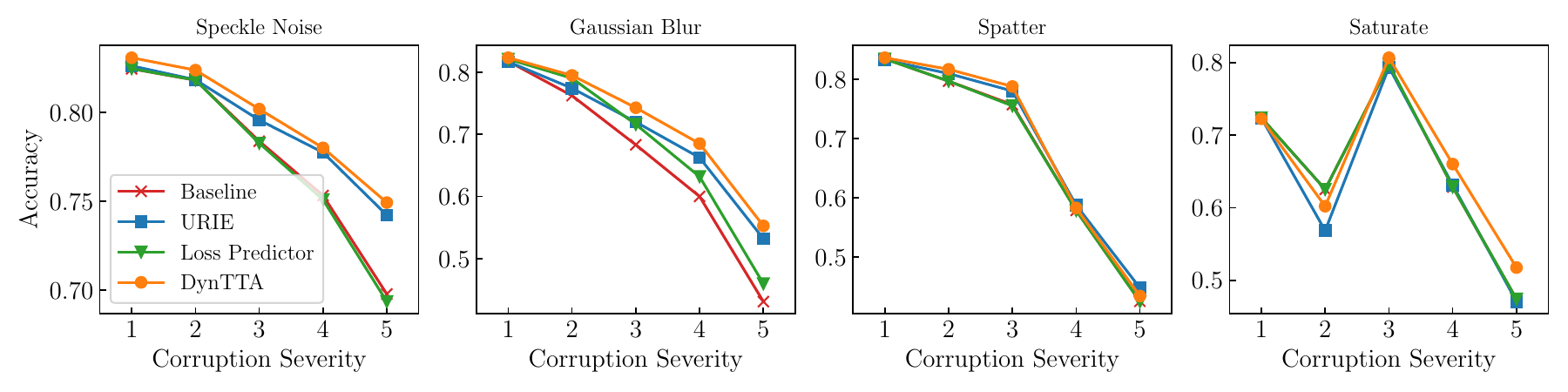}
\caption{Classification accuracy DeiT w/AugMix for each corruption and severity on the CUB dataset.}
\label{fig:nonblind_deit_augmix}
\end{figure*}

In this section, we show the accuracy for each corruption and severity results of the CUB dataset experiment presented in Section 4.2.
The results of the non-blind setting are shown in \Cref{fig:nonblind_r50}-\ref{fig:nonblind_deit_augmix}.
For \textit{Gaussian Noise} and \textit{Gaussian Blur}, DynTTA improves accuracy over comparison methods regardless of severity.
For \textit{Spatter}, DynTTA has about the same improvement in accuracy as the comparison method.
For \textit{Saturate}, DynTTA improves accuracy at higher severity compared to the comparison methods.

\begin{figure*}[tb!]
\centering
\includegraphics[width=1.0\linewidth]{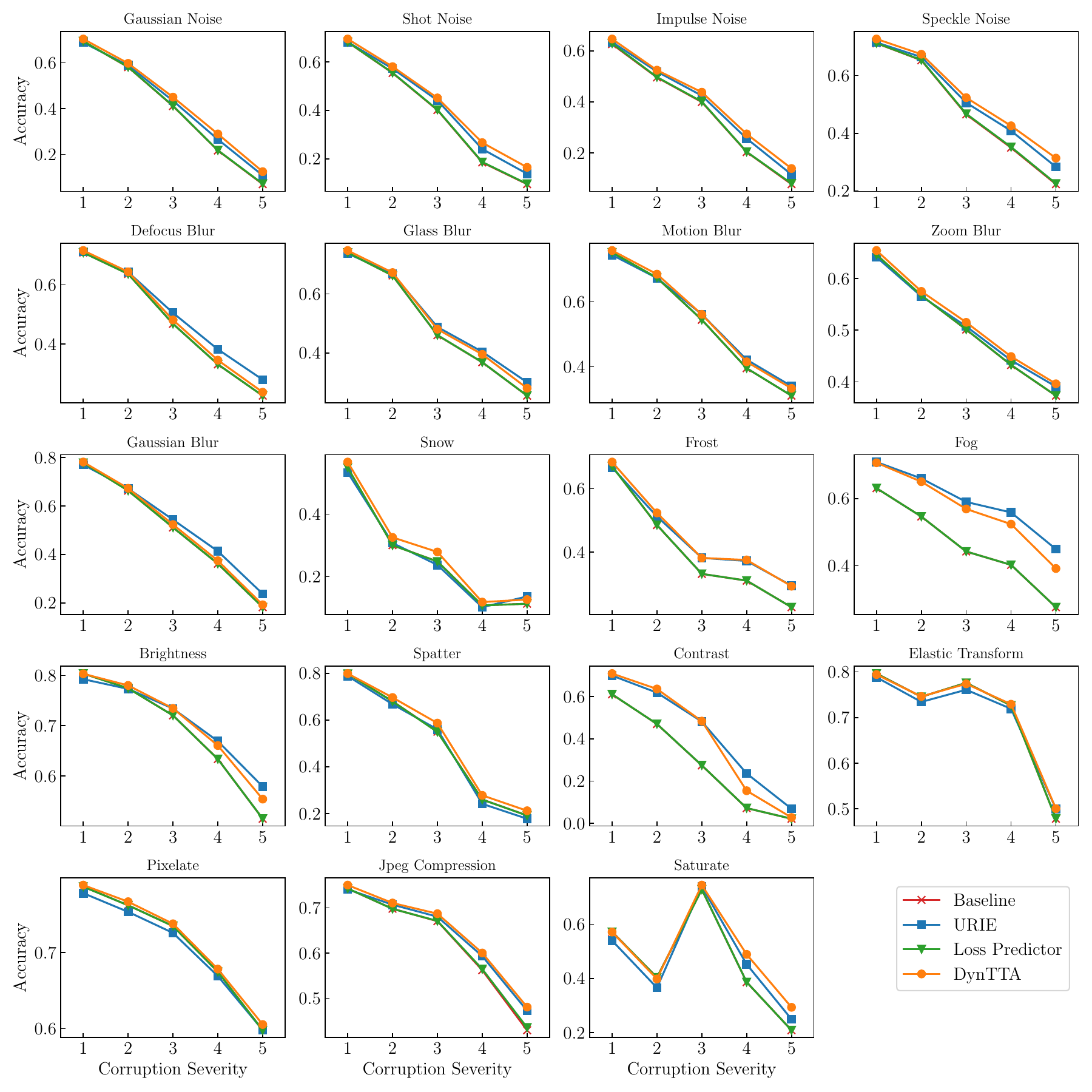}
\caption{Classification accuracy of ResNet50 for each corruption and severity on the CUB dataset.}
\label{fig:blind_r50}
\end{figure*}

\begin{figure*}[tb!]
\centering
\includegraphics[width=1.0\linewidth]{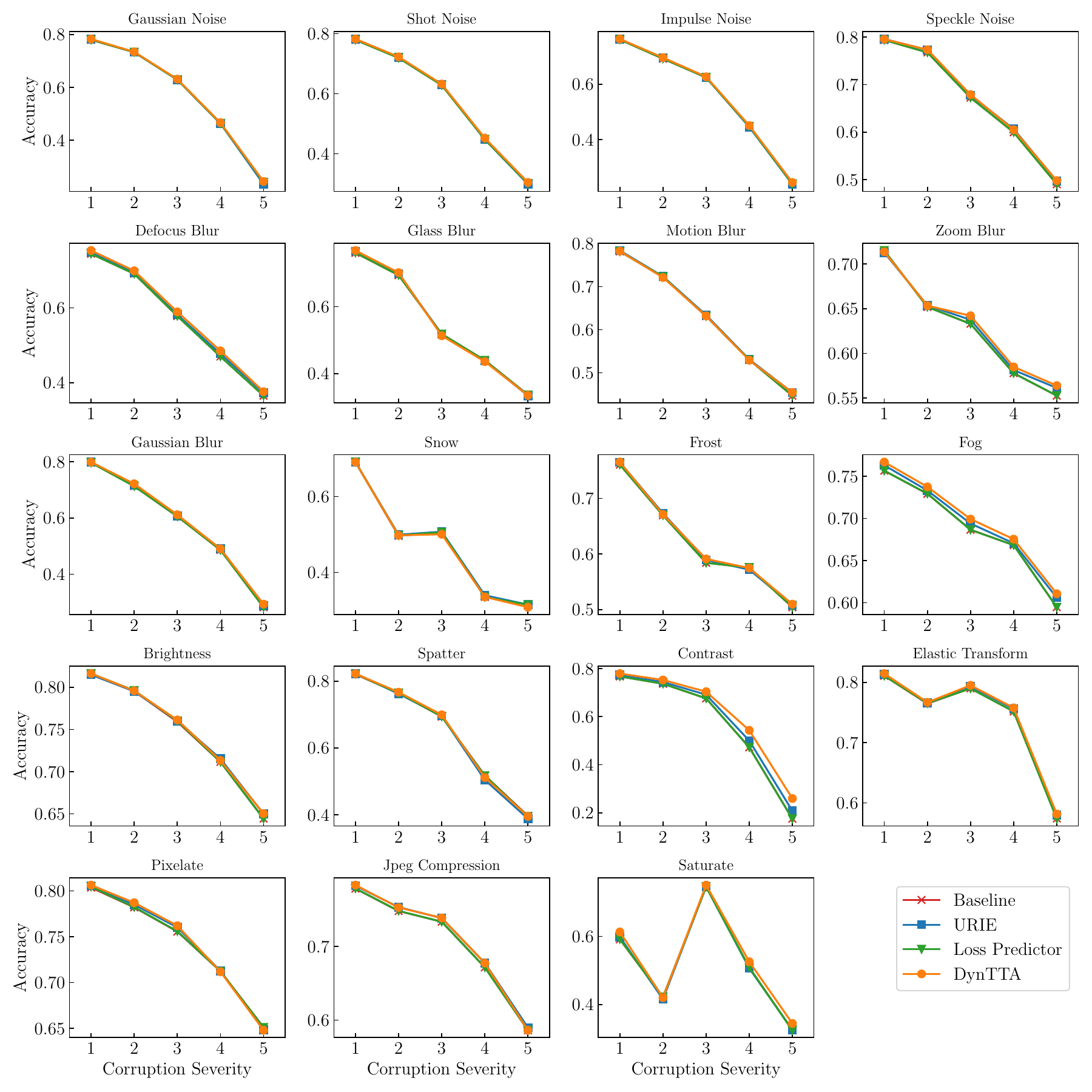}
\caption{Classification accuracy of ResNet50 w/AugMix for each corruption and severity on the CUB dataset.}
\label{fig:blind_r50_augmix}
\end{figure*}

\begin{figure*}[tb!]
\centering
\includegraphics[width=1.0\linewidth]{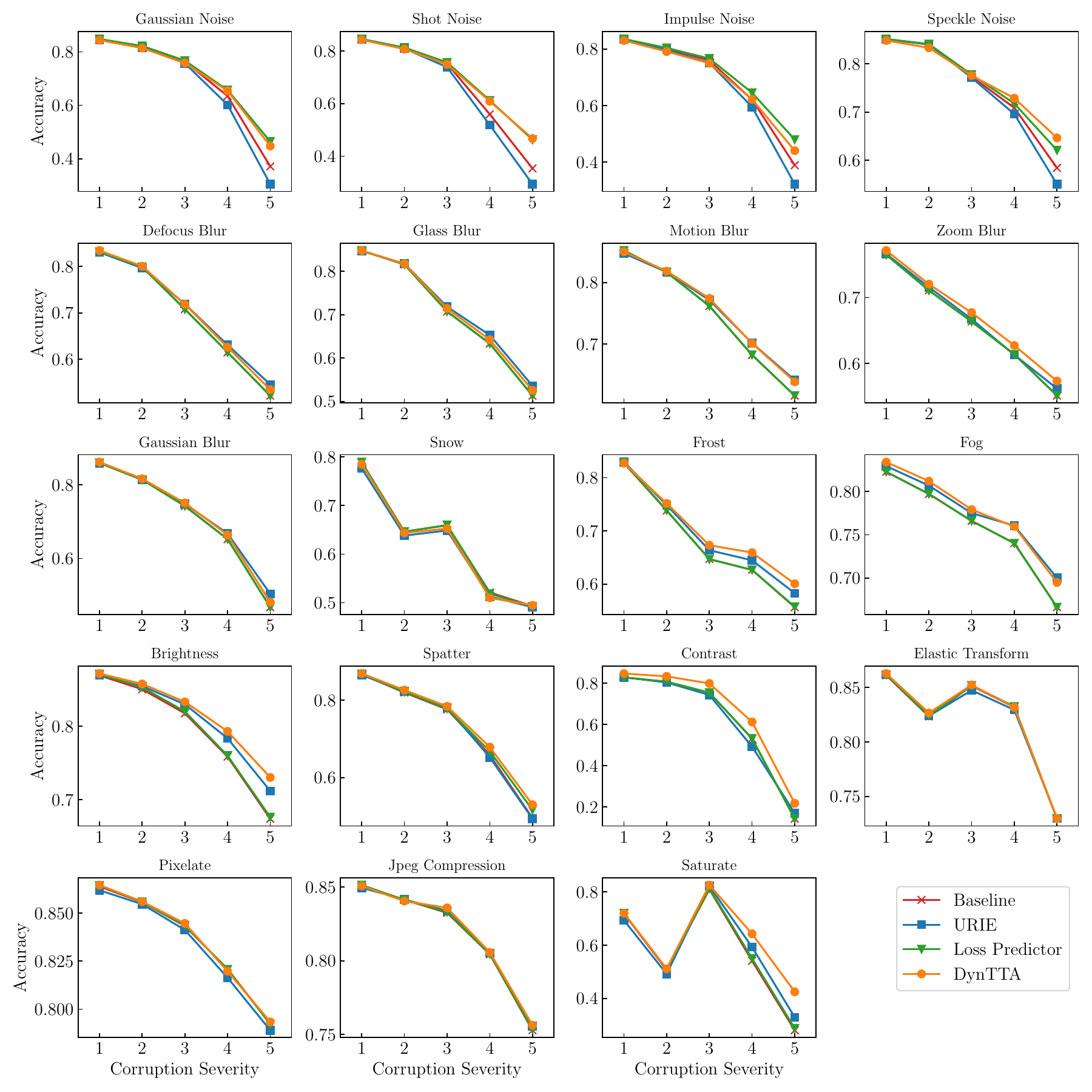}
\caption{Classification accuracy of Mixer-B16 for each corruption and severity on the CUB dataset.}
\label{fig:blind_mixer}
\end{figure*}

\begin{figure*}[tb!]
\centering
\includegraphics[width=1.0\linewidth]{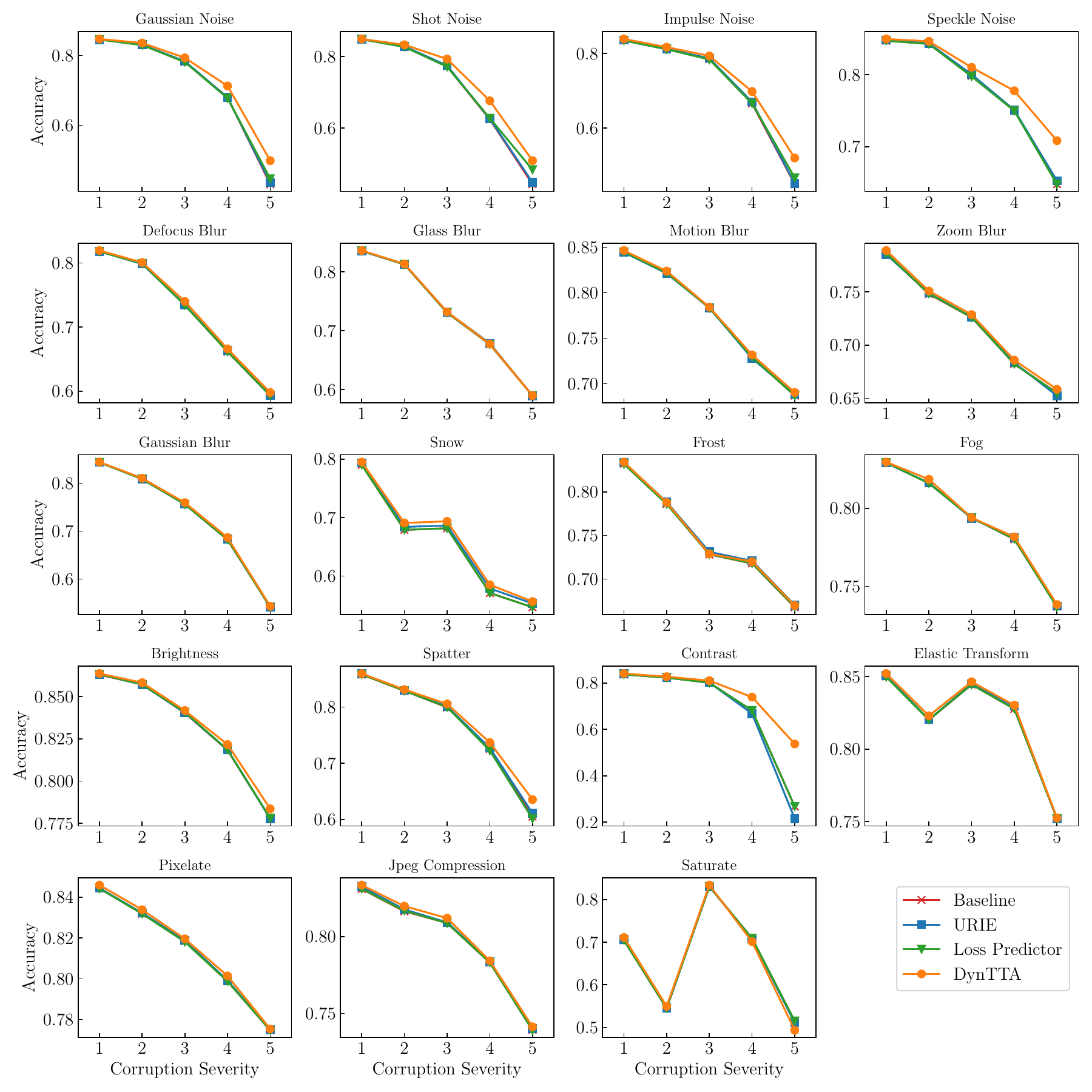}
\caption{Classification accuracy of Mixer-B16 w/AugMix for each corruption and severity on the CUB dataset.}
\label{fig:blind_mixer_augmix}
\end{figure*}

\begin{figure*}[tb!]
\centering
\includegraphics[width=1.0\linewidth]{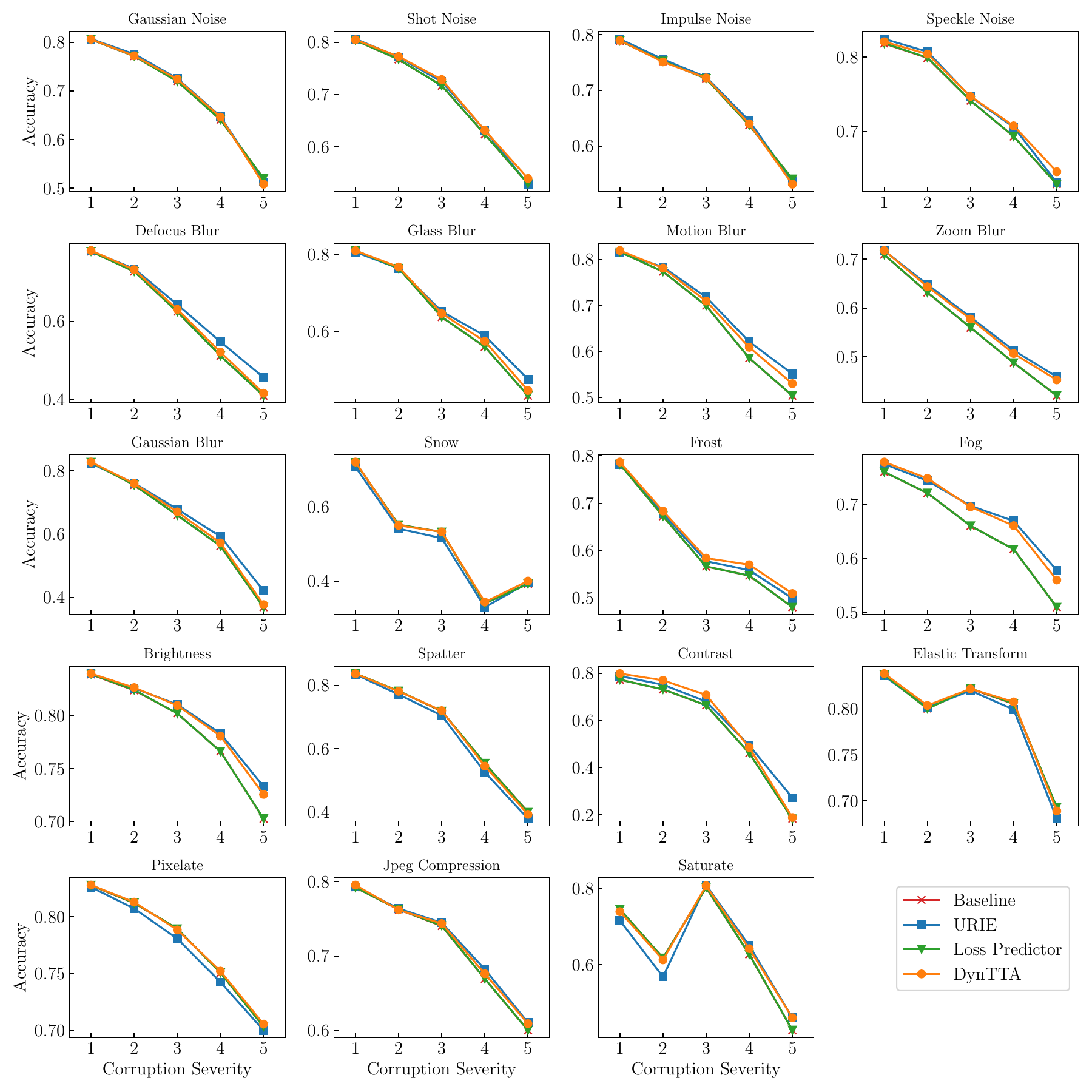}
\caption{Classification accuracy of DeiT for each corruption and severity on the CUB dataset.}
\label{fig:blind_deit}
\end{figure*}

\begin{figure*}[tb!]
\centering
\includegraphics[width=1.0\linewidth]{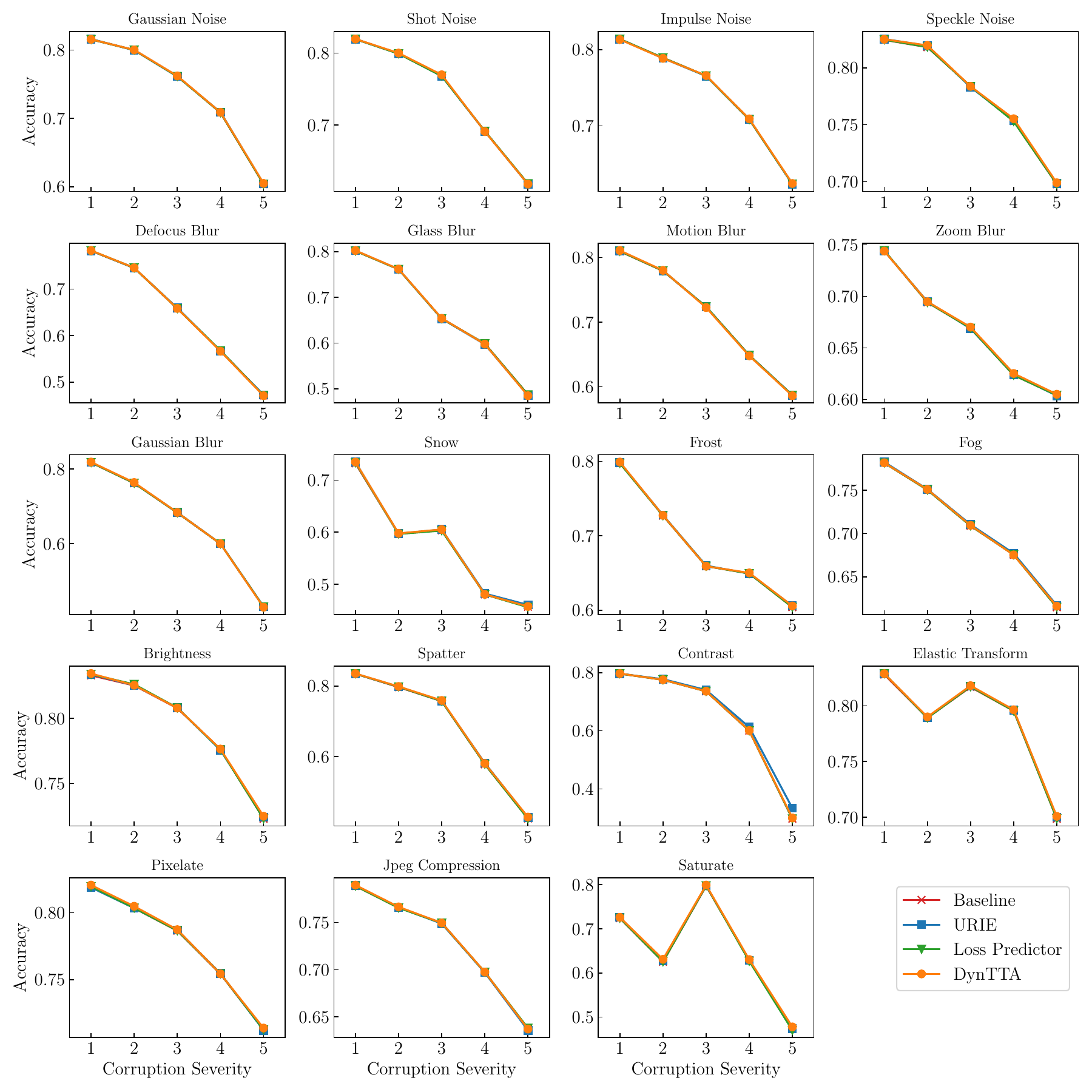}
\caption{Classification accuracy of DeiT w/AugMix for each corruption and severity on the CUB dataset.}
\label{fig:blind_deit_augmix}
\end{figure*}

The results of the blind setting are shown in \Cref{fig:blind_r50}-\ref{fig:blind_deit_augmix}.
Several corruptions are more difficult for Loss Predictor to improve compared to other methods due to its small augmentation space.
DynTTA outperforms the comparison methods in most experiments.
In particular, in experiments where the classification model is Mixer-B16 w/AugMix and the corruptions are \textit{Speckle Noise} and \textit{Contrast}, DynTTA improves accuracy while other methods do not.

\clearpage

\bibliographystyle{plain}
\bibliography{main}

\EOD